# DeepFEAv2: Deep Learning for Transient Finite Element Analysis Beyond Structured Meshes

Georgios Triantafyllou[1], Panagiotis G. Kalozoumis[1], Dimitris K. Iakovidis[1, *]
{gtriantafyllou, pkalozoumis, diakovidis}@uth.gr

[1] Department of Computer Science and Biomedical Informatics, University of Thessaly, 2-4 Papasiopoulou st. 35131, Lamia, Greece

**Abstract**

Finite Element Analysis (FEA) is widely used for transient mechanical simulations, but its high computational cost limits real-time and high-resolution applications. Deep learning surrogate models can reduce this cost; however, many existing approaches are restricted to steady-state prediction or cannot jointly predict Node- and Element-based Outputs (NEO) over time. The state-of-the-art DeepFEA framework has addressed these issues but remains limited to structured finite element (FE) meshes. To overcome this limitation, this study proposes DeepFEAv2, a deep learning surrogate framework that enables prediction of transient FEA simulations across different FE mesh topologies and element types. The main contributions of DeepFEAv2 are: (a) a module that uses the FE connectivity matrix to organize input features by element and arrange them into an input sequence guided by the mesh topology; (b) a novel neural network architecture designed to process the input sequence and jointly predict NEO over time; and (c) a FEA-informed optimization strategy for regularizing these NEO predictions. DeepFEAv2 was evaluated on structured and unstructured 3D linear elastic datasets, as well as on a pressure-driven aortic valve dataset. DeepFEAv2 achieved $R^2$ values up to 0.99 and normalized errors as low as 0.38%. Compared with DeepFEA, it achieved up to 38.0% relative increase in $R^2$ and up to 87.1% reduction in normalized error. DeepFEAv2 also performed inference up to three orders of magnitude faster than traditional FEA. These results demonstrate that DeepFEAv2 can efficiently model transient FEA simulations across increasingly complex FE settings, providing a scalable surrogate framework for transient FEA.

# 1 Introduction

Physical phenomena occurring in the natural world are predominantly governed by systems of Partial Differential Equations (PDEs) that encapsulate the spatiotemporal evolution of physical quantities (Evans, 2022). In research fields such as mechanical and biomedical engineering, Finite Element Analysis (FEA) serves as the primary computational framework for obtaining numerical solutions to these equations (Arzani et al., 2022; Samadian et al., 2025). FEA approximates the solution of PDEs by discretizing a continuous physical domain into a finite element (FE) mesh. A FE mesh consists of nodes and elements, where nodes represent discrete spatial points and elements define local regions of the physical domain by connecting groups of nodes (Ŝolín, 2005). The FE connectivity pattern between nodes and elements defines the mesh topology. In structured meshes, this topology follows a regular grid-like arrangement, whereas in unstructured meshes, it is described explicitly through irregular FE connectivity (Lo, 2014). In addition, FEA models may use different element types depending on the physical domain being represented; for example, solid elements discretize volumetric structures, whereas shell elements represent thin-walled structures using surface-like elements with an assigned thickness (Bathe, 2006). Using the resulting FE mesh, the governing PDEs are transformed into a finite-dimensional system of equations, which is then solved numerically, often through iterative procedures. In transient FEA, this numerical solution process is repeated over successive timesteps, where the solution at each timestep depends on the previous state of the simulated structure. While highly accurate, traditional FEA involves significant computational overhead that scales non-linearly with the order of the PDEs and the density of the mesh (Hughes, 2012). Consequently, the high computational cost of transient FEA creates a critical bottleneck for real-time engineering applications and the estimation of high-resolution solutions. This motivates the development of surrogate models that can approximate FEA simulations with substantially lower inference time.

[*] Corresponding author

To mitigate these computational challenges, Reduced-Order Modeling (ROM) and surrogate models have been extensively explored (Masoumi-Verki et al., 2022). Traditional ROM techniques, such as proper orthogonal decomposition and principal component analysis (Xiang et al., 2023), are frequently employed to reduce dimensionality but are fundamentally constrained by their linear nature. This has led to a paradigm shift toward deep learning-based surrogate models utilizing Artificial Neural Networks (ANNs) to approximate non-linear physical models (Du et al., 2022; Liang et al., 2020). Early steady-state surrogates primarily utilized Multilayer Perceptrons (MLPs) (Du et al., 2022) or Convolutional Neural Networks (CNNs) (Pfeiffer et al., 2019), but these methods lack the temporal awareness required to model scenarios where all states of a simulation must be predicted. Furthermore, a significant drawback of MLPs is their lack of spatial awareness regarding feature extraction (Farajtabar et al., 2023; Liu et al., 2022; Pellicer-Valero et al., 2020). In contrast, CNNs inherently offer spatial awareness and have been applied to predict deformation, stress, and strain (Ibragimova et al., 2022; Mendizabal et al., 2020; Pfeiffer et al., 2019). The complexity of transient FEA simulations requires models capable of capturing both spatial and temporal dependencies. Physics-Informed Neural Networks (PINNs) (Raissi et al., 2019) and Implicit Neural Representations (INRs) (Chen et al., 2023) have emerged as robust methods incorporating governing PDEs into their training loss function. Nevertheless, most physics-informed methods are PDE-specific and struggle to handle dynamic input predictions (Bolandi et al., 2023). For temporal analysis, recurrent architectures such as LSTMs (Chen et al., 2021) and GRUs (Wu et al., 2020) have been implemented, though they frequently lack spatial context. Spatiotemporal approaches, including Graph Neural Networks (GNNs) (Maurizi et al., 2022), CNN-LSTMs (Wang et al., 2021), and Convolutional LSTMs (ConvLSTMs) (Yan et al., 2023), have attempted to bridge this gap. Despite these advancements, existing transient models are often limited to 2D domains, rely on extensive ground truth data during inference, or fail to simultaneously predict distinct outputs for both the nodes and elements of FEA models (Triantafyllou et al., 2025). In addition, recursive models are highly susceptible to error propagation and accumulation over time (Franke and Wagner, 2024).

DeepFEA (Triantafyllou et al., 2025) was recently proposed as a spatiotemporal surrogate model for transient FEA simulations in order to overcome the aforementioned limitations. By leveraging a fused architecture of multilayer ConvLSTMs and parallel CNN branches, DeepFEA enabled the simultaneous prediction of node-based (displacement) and element-based (stress and strain) outputs (Triantafyllou et al., 2025). However, DeepFEA relies on regular grid-based mesh inputs, thus limiting its application to complex 3D unstructured meshes.

To address this limitation, which to the best of our knowledge is not addressed by any other state-of-the-art Deep Learning (DL)-based methodology, this study proposes a novel autoencoding DL framework designed to bridge the gap between surrogate modeling of transient FEA and unstructured FE meshes. DeepFEAv2 addresses the structured-mesh limitation of DeepFEA through two key innovations. First, it reorganizes nodal mesh features according to their FEs, without converting the physical domain into a regular grid-based input representation. The result of this reorganization is a one-dimensional input sequence, hereinafter referred to as an element sequence, where each entry corresponds to one FE and contains the features of its associated nodes. Second, DeepFEAv2 introduces a novel neural network designed to process the resulting element sequence using an autoencoding scheme that reduces computational burden, while producing output predictions for all nodes and elements of the FE mesh. As such, DeepFEAv2 is not constrained by the spatial layout, ordering, or size of the FE mesh and can therefore process different FE mesh organizations without requiring a regular grid-based input representation. In summary, the innovative contributions of this study include:

- A deep learning surrogate framework for transient FEA: This study proposes DeepFEAv2, a deep learning surrogate framework designed to predict transient FEA responses across different FE mesh topologies and element types. In contrast to CNN- or ConvLSTM-based surrogate models that require a regular grid-based input representation, DeepFEAv2 reorganizes FE mesh data into an element sequence through a mesh connectivity-based grouping (MCBG) and adjacency-based element ordering (ABEO) module. As a result, structured and unstructured FE meshes can be seamlessly processed within the same modelling framework.
- Topology-aware node-element prediction network: This study introduces a novel neural network composed of a one-dimensional CNN autoencoder, followed by 3D ConvLSTM layers and two decoder branches. Unlike other surrogate models, such as DeepFEA, that model temporal evolution directly over the full mesh representation, DeepFEAv2 performs temporal prediction in a learned reduced-order space. This design lowers the computational burden, enabling DeepFEAv2 to operate on large meshes while maintaining a spatially organized representation of the compressed FEA features.
- FEA-informed node-element loss optimization: This study extends the previous node-element loss formulation of DeepFEA with additional geometry- and time-consistency terms. Compared with purely data-driven loss functions that optimize only nodal or element-based output errors, the proposed objective jointly optimizes nodal displacement, effective stress, and effective strain while penalizing nonphysical edge-length changes, surface-normal inconsistencies and temporal consistency errors.

- Evaluation across different mesh topologies and element types: This study validates DeepFEAv2 on structured and unstructured solid-element 3D Linear Elastic Model (LEM) datasets and a pressure-driven aortic valve simulation dataset based on shell elements, demonstrating the robustness of the proposed framework across different mesh organizations, element formulations, loading mechanisms, and geometric complexities.

To the best of our knowledge, DeepFEAv2 is the first transient FEA surrogate framework to combine MCBG, autoencoding-based dimensionality reduction, latent-space 3D ConvLSTM temporal modeling, and FEA-informed node-element loss optimization for transient FEA simulations across different mesh topologies.

The remainder of this paper is organized as follows: Section 2 reviews the related work on deep learning-based FEA surrogate modeling, mesh-based learning, and spatiotemporal predictions; Section 3 presents the proposed methodology, including the input and output tensor definitions, the MCBG and ABEO preprocessing modules, the Topology-aware Node-Element Prediction (TopoNEP) network and the training loss formulation; Section 4 describes the datasets, evaluation metrics, experimental setup, results for the structured and unstructured 3D LEM cases, presents an application study on a pressure-driven aortic valve simulation to evaluate the proposed framework on a complex shell-based geometry and reports the inference time analysis and compares the computational cost of DeepFEAv2 with conventional FEA and DeepFEA; Section 5 discusses the main findings, model behavior, advantages, and limitations; Section 6 summarizes the conclusions and outlines future research directions.

# 2 Related Work

The development of AI-based surrogate models to accelerate or replace computational FEA solvers has seen significant interest, generally falling into three categories: steady-state, physics-informed, and transient analysis models. Early steady-state approaches utilized MLPs either to assist traditional solvers (Bender et al., 2019; Burghardt et al., 2021) or to completely bypass them by directly mapping parameters to outputs (Du et al., 2022; Farajtabar et al., 2023; Pellicer-Valero et al., 2020). While implementation is straightforward, MLPs inherently lack spatial awareness and become computationally prohibitive when scaled to large meshes. To capture the spatial hierarchies of physical models, CNN-based architectures, including Bayesian CNNs and U-Net variants, have been adopted (Deshpande et al., 2022; Krokos et al., 2022; Mendizabal et al., 2020). Although CNNs scale efficiently and inherently provide spatial awareness, they are heavily reliant on structured grids and lack the temporal mechanisms required to predict sequential states in dynamic simulations.

Physics-informed methods have also been widely explored as a means to approximate FEA simulation predictions by incorporating physical laws directly into the training process of the network. PINNs and INRs evaluate spatial and temporal variables by embedding the governing PDEs and boundary conditions directly into the loss function (Bolandi et al., 2023; Chen et al., 2023; Haghighat et al., 2021; Haubner et al., 2024). These approaches are highly effective for interpolation. However, their training is rigidly tailored to specific PDEs, and they are generally inflexible when applied to transient scenarios that feature highly dynamic, time-varying input conditions.

For transient FEA simulations, models must successfully capture spatiotemporal dependencies and mitigate the propagation of error across iterative predictions. Standard recurrent models, such as LSTMs and GRUs, capture temporal dynamics but lack spatial context (Chen, 2021; Triantafyllou et al., 2025). Consequently, recent literature has shifted toward hybrid spatiotemporal architectures. Fusions of CNNs and LSTMs have been utilized for Fluid-Structure Interaction (FSI) and single-parameter predictions, though they remain largely constrained to 2D domains (Arcones et al., 2022; Chijioke et al., 2022). GNNs offer a compelling alternative by treating meshes as graphs to maintain spatial awareness, but they face scalability issues and struggle with the dynamic graph characteristics required for long-term transient inference (Maurizi et al., 2022; Sharma et al., 2024). While ConvLSTMs have been successfully applied to 2D microstructure evolution (Frankel et al., 2020) and material design (Yan et al., 2023), current transient surrogate models collectively lack the capacity to seamlessly handle dynamic inputs, support 3D topologies, and simultaneously predict multi-dimensional element and node outputs without relying on extensive initial ground truth data. To overcome these spatiotemporal challenges, DeepFEA was introduced as the current state-of-the-art for transient FEA. By utilizing a fused architecture of multilayer ConvLSTM and parallel CNN branches, DeepFEA successfully demonstrated simultaneous node-level (displacement) and element-level (stress and strain) output predictions. Nevertheless, a significant limitation of the DeepFEA framework is that it is not applicable to unstructured meshes.

To address this limitation, this study introduces DeepFEAv2, an extension of the previous DeepFEA framework designed to support different FE mesh topologies. Unlike current deep learning-based FEA surrogate approaches, which are often limited by structured-grid requirements, mesh-specific formulations, static prediction settings, or the prediction of only a single response type, DeepFEAv2 utilizes the FE connectivity matrix to model transient evolution across arbitrary structured and unstructured 3D FE topologies. Specifically, DeepFEAv2 overcomes prior limitations by organizing mesh input features into a topology-aware element sequence guided by the FE connectivity matrix, coupled with an autoencoding scheme, a neural network mechanism that compresses high-dimensional mesh data into a compact set of learned features, to model temporal dynamics within a reduced-order representation and jointly predict node- and element-based outputs (NEO). By resolving the conflict between varying mesh topologies and convolutional feature extraction, DeepFEAv2 provides a versatile and scalable surrogate model capable of replacing traditional FEA solvers in high-resolution domains for time-efficient computation.

# 3 Methods

DeepFEAv2 is a deep learning framework designed to predict the output of transient FEA simulations over multiple timesteps. Figure 1 provides an overview of DeepFEAv2 and illustrates the complete data flow from the simulation inputs to the predicted NEO. The core novelty of DeepFEAv2 is the integration of MCBG, which groups nodal features by element using mesh connectivity, ABEO, which arranges elements into a 1D sequence based on physical adjacency, and a TopoNEP network, which compresses, temporally evolves, and decodes joint NEO predictions, into a single transient FEA surrogate framework. The workflow consists of four main stages, following the numbering shown in Fig. 1.

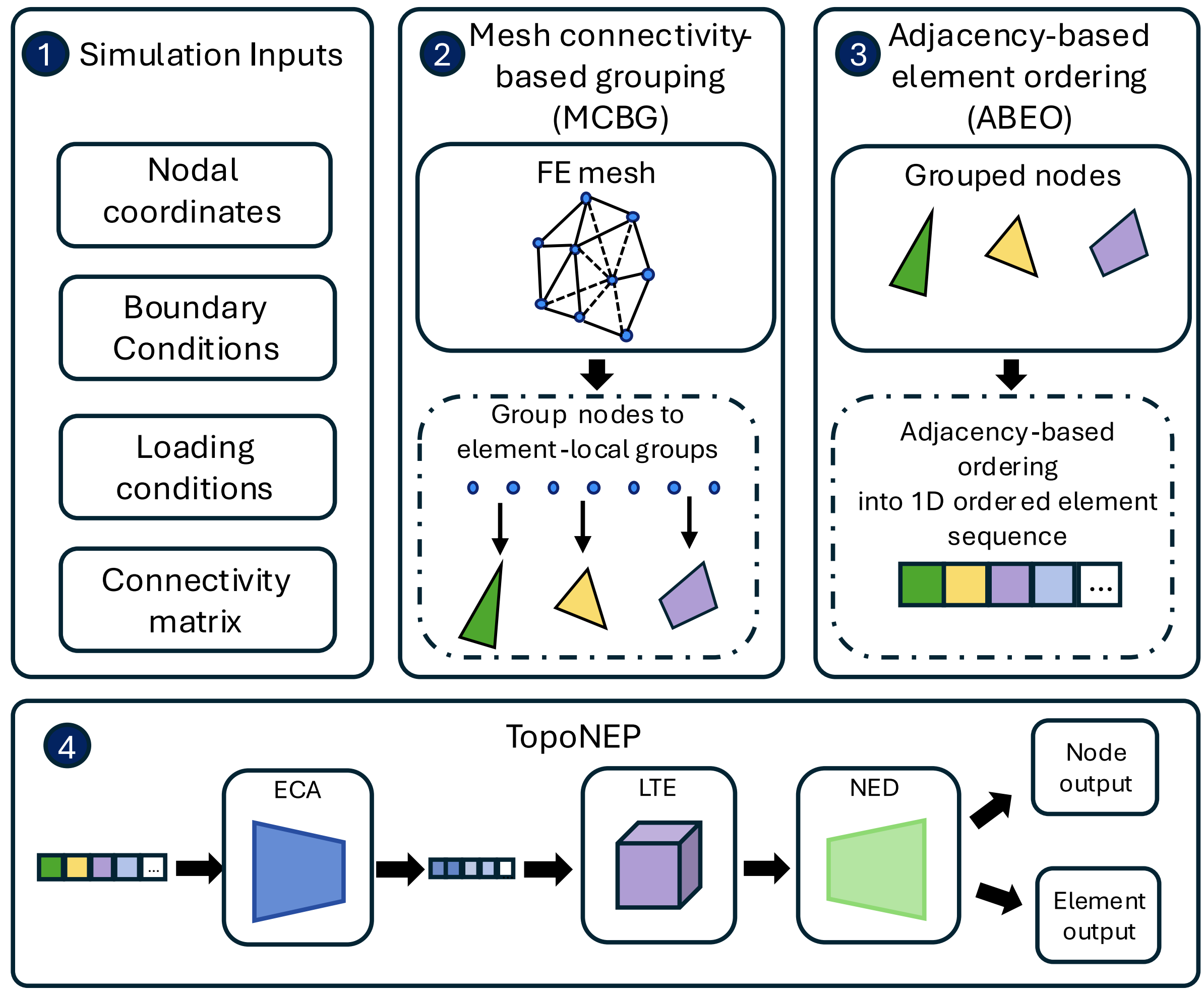

*Figure 1 Overview of the proposed framework that utilizes a Topology-aware Node-Element Prediction (TopoNEP) network, comprising an Element-Centric Autoencoder (ECA), a Latent Temporal Evolution (LTE) module, and a Node-Element Decoder (NED), to generate node- and element-based outputs.*

First, the simulation inputs are defined, including the nodal coordinates, boundary conditions, loading conditions, e.g., applied force, and the FE connectivity matrix, *i.e.*, the matrix which specifies the global node indices associated with each finite element. These inputs provide the geometric, physical, and topological information required by the framework. Second, MCBG is applied to the FE mesh. In this stage, the FE connectivity matrix is used to group nodal features into element-local groups, where each group contains the features of the nodes belonging to a specific element. Third, ABEO is performed to arrange the element-local groups into an ordered one-dimensional element sequence. This ordering is based on element adjacency, so that neighboring elements in the original FE mesh are placed close to one another in the element sequence. Fourth, the ordered element sequence is passed to TopoNEP. TopoNEP consists of three main modules, namely the Element-Centric Autoencoder (ECA), the Latent Temporal Evolution (LTE) module, and the Node-Element Decoder (NED). The ECA compresses the element sequence into a compact latent representation. This module is termed element-centric because each input entry to the autoencoder represents one FE, which is formed by grouping the features of the nodes that belong to that element. The LTE module then models the transient evolution of this representation in a compact 3D latent space using sequential 3D ConvLSTM layers. In this context, the latent space denotes the reduced-order representation of the high-dimensional FE mesh state produced by the ECA, where the ordered element sequence is compressed into a compact set of learned features. Finally, the NED maps the evolved latent features into the physical FE mesh domain through two CNN-based prediction branches, *i.e.*, a node branch for node output and element branch for element output.

TopoNEP is trained based on an expanded version of the Node-Element Loss Optimization (NELO) algorithm, originally introduced in (Triantafyllou et al., 2025). This optimization algorithm gradually exposes the network to its own predictions to reduce error accumulation produced by recursive predictions of the network. Accordingly, the previous node-element loss is expanded with FE-informed geometry- and time-consistency terms resulting in a FEA-informed Node-Element loss $L_{\text{FNE}}$. During inference, the trained TopoNEP recursively predicts the entire output sequence of the simulation based only on the initial mesh state, boundary conditions, connectivity information and loading conditions. The following subsections describe the input and output tensor formulation, the preprocessing pipeline using MCBG and ABEO, TopoNEP, and the training procedure.

## 3.1 Input and Output Tensors

The geometry of the structure to be analyzed is first discretized into an FE mesh composed of nodes and elements. Nodes correspond to discrete spatial points, while elements define local regions of the mesh by connecting predefined groups of nodes. In structured FE meshes, the node–element connectivity follows a grid-like organization and can therefore be encoded directly as a tensor. In unstructured FE meshes, the data can still be stored as tensors, but the node–element connectivity is not implied by tensor indices and must be specified explicitly by the FE connectivity matrix.

The input tensor always contains two fundamental components, *i.e.*, the current nodal coordinates of the mesh and a boundary condition map indicating which nodes are constrained. All remaining input features depend on the type of the FEA simulation model. At timestep $t$, the general input tensor is defined as $X_t = (N_t, S_t, \Gamma) \in \mathbb{R}^{B \times F_{\text{in}} \times \Lambda}$, where $N_t$ denotes the current nodal coordinates, $\Gamma$ denotes the boundary-condition map, and $S_t$ denotes the set of simulation-specific input features, $B$ is the batch size, $F_{\text{in}}$ is the total number of input feature channels and $\Lambda$ are the total number of nodes. For a 3D mesh, $N_t = (N_x^t, N_y^t, N_z^t)$, where $N_x^t$, $N_y^t$, and $N_z^t$ are the x-, y-and z- axis coordinate tensors respectively, of all mesh nodes at timestep $t$. The $\Gamma$ map is represented as a binary nodal feature, where constrained nodes are assigned 0 and free nodes are assigned 1. This feature enables the network to distinguish between fixed and deformable regions of the mesh. The simulation-specific components $S_t$ depend on the examined FEA problem. For example, in a force-driven structural simulation, $S_t$ may include the externally applied nodal force components, $S_t = (F_x^t, F_y^t, F_z^t)$; in a pressure-driven simulation, $S_t$ may contain pressure-related descriptors; in a thermomechanical simulation, it may include temperature or thermal-gradient fields. Therefore, the general input tensor for a 3D simulation can be expressed as $X_t = (N_x^t, N_y^t, N_z^t, \Gamma, S_1^t, S_2^t, \dots, S_m^t)$, where $m$ is the number of simulation-specific input feature channels.

Finally, the output is divided into two tensors, *i.e.*, a node output tensor, $\hat{Y}_{t+1}^n$, and an element output tensor, $\hat{Y}_{t+1}^e$. In this study, the node output tensor represents the nodal displacement, $(U_x^{t+1}, U_y^{t+1}, U_z^{t+1})$, where $U_x^{t+1}$, $U_y^{t+1}$, and $U_z^{t+1}$ denote the displacement components of each node along the $x$-, $y$-, and $z$-axis, respectively. The element output tensor comprises effective stress, $\Sigma_{t+1}$, and effective strain, $E_{t+1}$. This distinction is necessary since different FEA quantities are naturally defined on different parts of the mesh. Nodal coordinates and displacements are associated with nodes, whereas stress and strain are associated with elements.

## 3.2 Mesh Connectivity-Based Grouping and Adjacency-Based Element Ordering

DeepFEAv2 utilizes Mesh Connectivity-Based Grouping (MCBG) and Adjacency-Based Element Ordering (ABEO) to transform raw FE simulation inputs into a topology-aware sequence specifically designed for processing by the TopoNEP network (Fig. 2). First, MCBG uses the FE connectivity matrix to organize raw nodal features into element-local representations. Next, ABEO arranges these element representations into a continuous 1D sequence based on physical mesh adjacency, ensuring that elements sharing edges or faces in the original mesh are positioned close to one another in the input sequence.

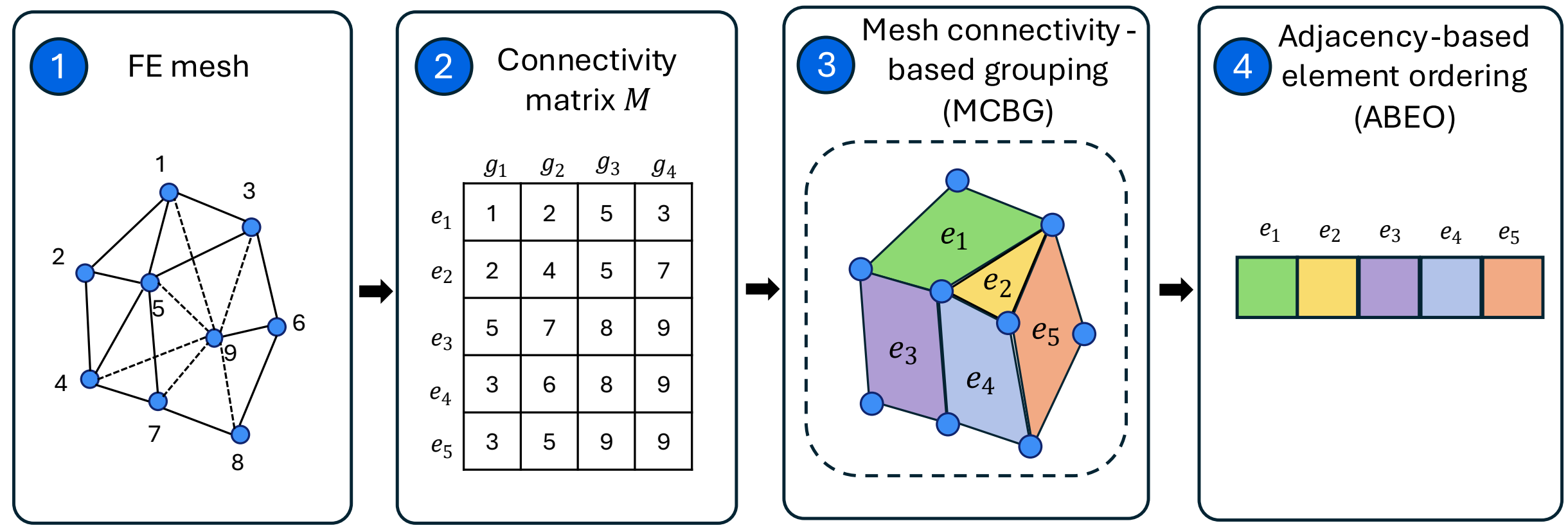

*Figure 2 Overview of the input preprocessing pipeline combining Mesh Connectivity-Based Grouping (MCBG) and Adjacency-Based Element Ordering (ABEO).*

Let the FE mesh contain $E$ elements. Since different element types may be defined by different numbers of nodes, the number of nodes of element $e$, hereinafter local nodes, is denoted by $G_e$. To represent all elements in a common tensor form, $G_{\max} = \max_e G_e$ is defined as the maximum number of local nodes among all elements in the mesh. The FE connectivity is then represented as $M \in \mathbb{Z}^{E \times G_{\max}}$, where $M[e, g]$ denotes the global node index assigned to the $g$-th local node of element $e$. For elements with fewer than $G_{\max}$ nodes, the empty cells in $M$ are filled by the latest used node index of the node in the latest non-empty cell from the same element row. For example, if we consider a FE connectivity matrix $M$ with $G_{\max} = 4$ and an element $e$ with $G_e = 3$, then $M[e, g] = [g_1, g_2, g_3, g_3]$, where $g_1$, $g_2$ and $g_3$ represent the global node indices of the three nodes associated with element $e$. This padding is used only to obtain a fixed tensor shape for batching and convolutional processing. A validity mask $V \in \{0,1\}^{E \times G_{\max}}$ is also defined, where $V[e, g] = 1$ for cells with physical nodes and $V[e, g] = 0$ for padded cells. Therefore, repeated padded entries are not treated as additional physical nodes and $G_e = \sum_{g=1}^{G_{\max}} V[e, g]$.

For each element, the corresponding nodal features are grouped according to $M$, producing the element-local representation:

$$X_{\text{local}}^{t}(b, e, :, g) = V[e, g] \cdot X_t(b, :, M[e, g]), g = 1, \dots, G_{\max} \quad (1)$$

where, $b$ indexes the batch sample, $g$ indexes the local node slot, and $X_{\text{local}}^{t} = (N_{\text{local}}^{t}, S_{\text{local}}^{t}, \Gamma_{\text{local}}) \in \mathbb{R}^{B \times E \times F_{\text{in}} \times G_{\max}}$ is the element-local representation of $X_t$. Similarly, $N_{\text{local}}^{t}$, $S_{\text{local}}^{t}$, and $\Gamma_{\text{local}}$ denote the element-local representations of $N_t$, $S_t$, and $\Gamma$, respectively. When operations depend on the physical local nodes of an element, such as nodal feature averaging, mesh-edge construction, or topology-based regularization, only entries with $V[e, g] = 1$ are used. Therefore, padded duplicate entries do not affect the physical connectivity or the normalization of local node operations.

Since the proposed encoder uses 1D convolution along the element dimension, the order of the elements is important. Therefore, the elements are arranged using ABEO (Fig. 2), where element centroids are first computed from the undeformed mesh coordinates, and an adjacency graph is constructed based on shared edges or faces. Then, the elements are traversed so that physically neighboring elements are placed close to one another in the element sequence.

## 3.3 Topology-Aware Node-Element Prediction Network Architecture

The TopoNEP architecture consists of three main modules: (i) an ECA module; (ii) a LTE module and (iii) a NED module. The overall purpose of the architecture is to compress the high-dimensional mesh representation, then model its temporal evolution in a compact latent space and finally decode this latent space into the NEO.

### 3.3.1 ECA: The Element-Centric Autoencoder Module

At each timestep $t$, the input tensor is first converted into $X_{\text{local}}^{t}$ as described in Section 3.2. The features of the $G_{\max}$ nodes that define each element are concatenated along the feature channel dimension resulting in a reshaped input tensor $X_{\text{enc}}^{t} \in \mathbb{R}^{B \times (F_{\text{in}} \cdot G_{\max}) \times E}$. This produces the input to the ECA module that is responsible for compressing $X_{\text{enc}}^{t}$ into a lower-dimensional latent representation. The ECA comprises a sequence of encoding blocks, where each block contains a 1D convolutional layer (Conv1D) followed by instance normalization and the nonlinear activation function, ReLU (Fig. 3). After each encoding block, average pooling is used to further reduce the length of the element sequence along the $E$ dimension. After the final encoding block, adaptive average pooling is used to map the encoded representation to a fixed latent size. The output is then reshaped into a 3D latent volume, $Z_t \in \mathbb{R}^{B \times C_{\text{lat}} \times H_{\text{lat}} \times W_{\text{lat}} \times D_{\text{lat}}}$, where $C_{\text{lat}}$ is the number of latent feature channels, and $H_{\text{lat}}$, $W_{\text{lat}}$, and $D_{\text{lat}}$ define the spatial dimensions of the compressed representation. Overall, the ECA module can be formally defined as follows:

$$Z_t = \mathcal{E}_\theta(X_{\text{enc}}^t) \tag{2}$$

where $\mathcal{E}_\theta$ denotes the trainable encoder.

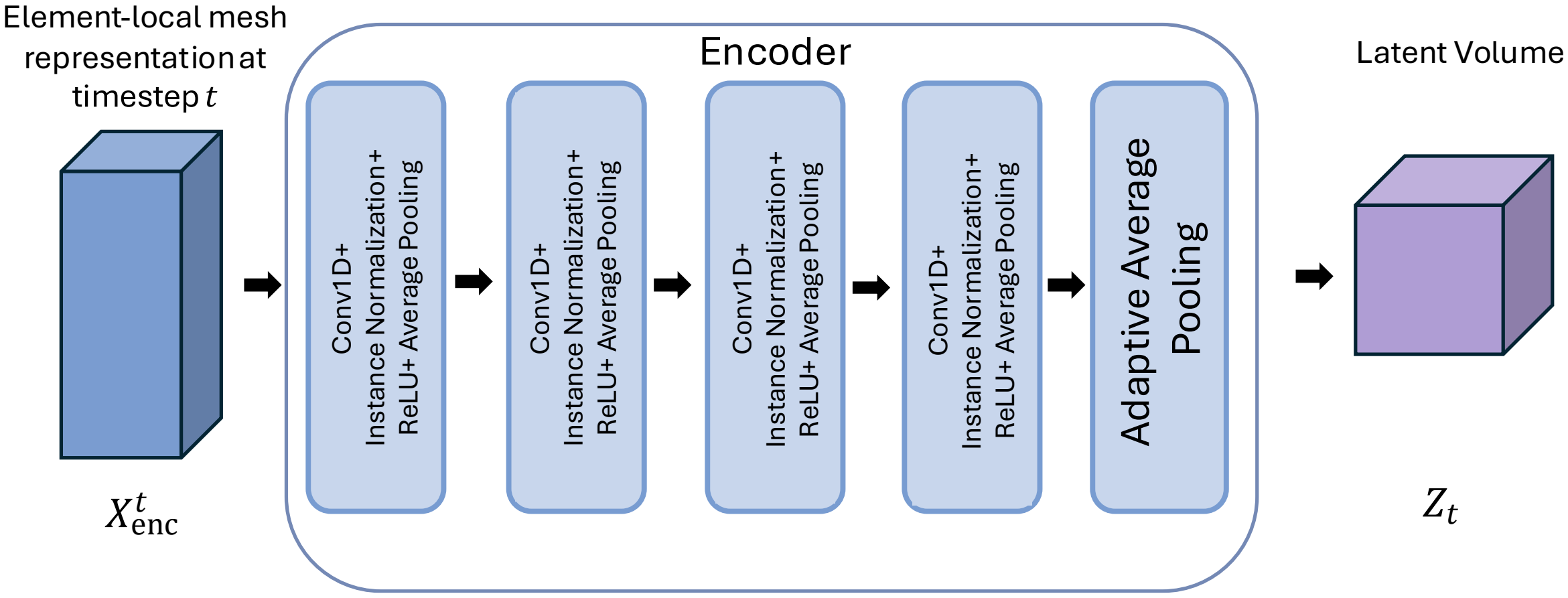

*Figure 3 Block diagram of the ECA module.*

It is important to note that the produced latent volume is not a regular grid-based representation of the original mesh. Instead, it is a learned compact manifold derived from the element sequence. Hence, the structured 3D organization is introduced only in the latent space and is not imposed on the entire physical FE mesh. Therefore, the MCBG, ABEO and ECA steps create a compact latent space in which efficient 3D recurrent operations can be performed. This autoencoding design provides two main advantages. First, it reduces the dimensionality of the original mesh representation before temporal modeling, lowering memory requirements and computational cost. Second, it enables the temporal dynamics of the system to be modeled in a spatially organized latent representation, allowing the recurrent module to preserve local structure within the compressed domain.

### 3.3.2 LTE: The Latent Temporal Evolution Module

The temporal evolution of the encoded simulation state is modeled by LTE, which consists of three 3D ConvLSTM layers with kernel size 3 (Fig. 4). The number of layers and the kernel size were selected based on (Triantafyllou et al., 2025). At timestep $t$, the LTE module receives the latent volume $Z_t$ produced by the ECA module. This latent volume is provided as input to the first ConvLSTM layer. Each ConvLSTM layer $l$ maintains two recurrent states, *i.e.*, a hidden state, ($H_t^l$), which represents the recurrent output state of the layer, and a cell state, ($C_t^l$), which represents its internal memory state. During timestep $t$, each layer updates its previous states, $H_{t-1}^l$ and $C_{t-1}^l$, using the current input to that layer. The hidden state $H_t^l$ is propagated both to the next ConvLSTM layer and to the same layer at the next timestep, while the cell state $C_t^l$ is propagated through time within the same layer.

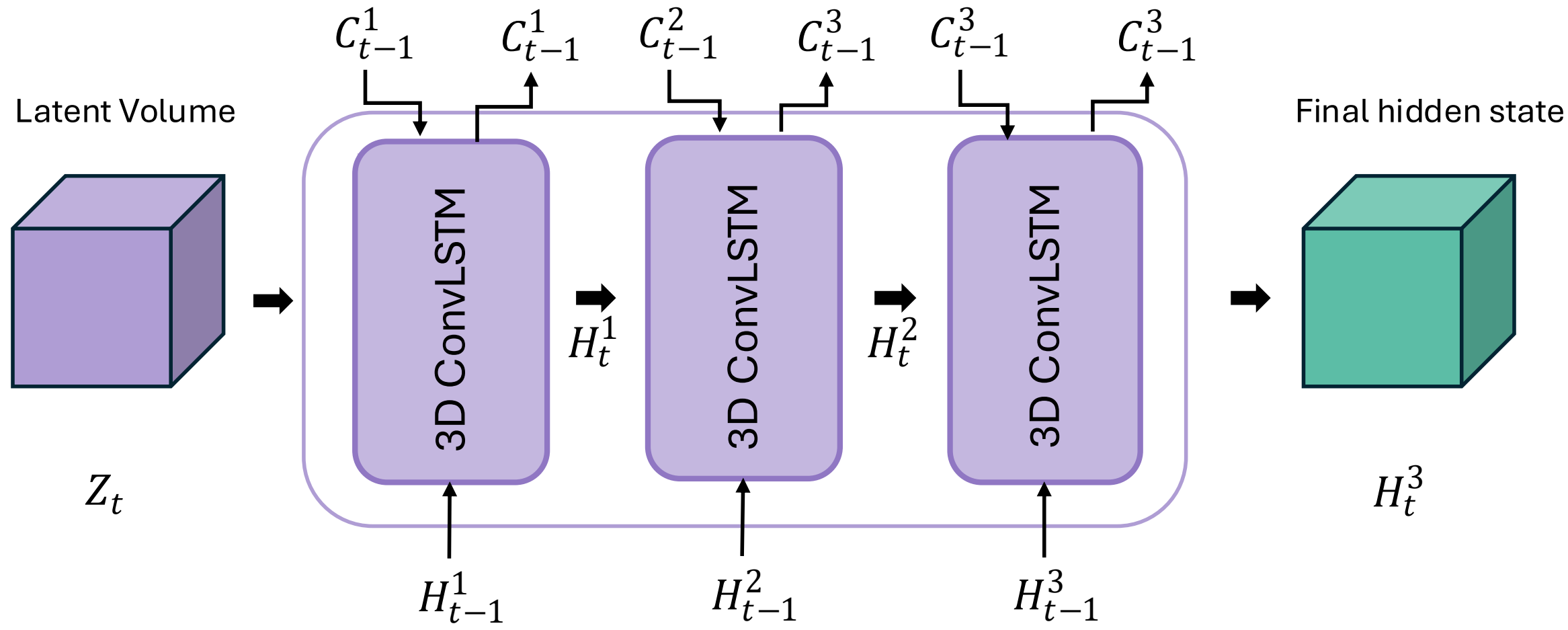

*Figure 4 Block diagram of the Latent Temporal Evolution Module.*

Let $Q_t^l$ denote the input to ConvLSTM layer $l$, where $Q_t^1 = Z_t$ for the first layer and $Q_t^l = H_t^{l-1}$ for $l > 1$. A 3D ConvLSTM layer can be formally defined as follows:

$$\varphi(\cdot\,; W, b) = \begin{cases} i_t^l = \sigma\left(W_{xi}^l * Q_t^l + W_{hi}^l * H_{t-1}^l + W_{ci}^l \odot C_{t-1}^l + b_i^l\right) \\ f_t^l = \sigma\left(W_{xf}^l * Q_t^l + W_{hf}^l * H_{t-1}^l + W_{cf}^l \odot C_{t-1}^l + b_f^l\right) \\ C_t^l = f_t^l \odot C_{t-1}^l + i_t^l \odot tanh\left(W_{xc}^l * Q_t^l + W_{hc}^l * H_{t-1}^l + b_c^l\right) \\ o_t^l = \sigma\left(W_{xo}^l * Q_t^l + W_{ho}^l * H_{t-1}^l + W_{co}^l \odot C_t^l + b_o^l\right) \\ H_t^l = o_t \odot tanh(C_t^l) \end{cases} \tag{3}$$

where $*$ denotes 3D convolution, $\odot$ denotes elementwise multiplication, $\sigma$ is the sigmoid function and $tanh$ is the hyperbolic tangent. The terms $i_t$, $f_t$, and $o_t$ denote the input, forget, and output gates, respectively. These gates represent mathematical operations that regulate information flow through the recurrent memory, *i.e.*, the input gate controls how much new information is added, the forget gate controls how much previous information is retained, and the output gate controls how much of the updated memory contributes to the hidden state. The tensors $W_{xi}^l$, $W_{xf}^l$, $W_{xc}^l$, and $W_{xo}^l$ are the convolutional kernels applied to the current layer input $Q_t^l$, whereas $W_{hi}^l$, $W_{hf}^l$, $W_{hc}^l$, and $W_{ho}^l$ are the convolutional kernels applied to the previous hidden state $H_{t-1}^l$. The terms $W_{ci}^l$, $W_{cf}^l$, and $W_{co}^l$ denote the weights associated with the previous cell state $C_{t-1}^l$. Lastly, $b_i^l$, $b_f^l$, $b_c^l$, and $b_o^l$ denote the bias terms of the input gate, forget gate, cell-state update, and output gate, respectively.

After the three stacked ConvLSTM layers are applied, the hidden state of the final layer, $H_t^3 \in \mathbb{R}^{B \times C_{\text{out}} \times H_{\text{lat}} \times W_{\text{lat}} \times D_{\text{lat}}}$, where $C_{\text{out}}$ denotes the output feature channel dimension of the final ConvLSTM layer, is used as the temporally evolved latent representation of the FE simulation state. This allows DeepFEAv2 to model transient evolution in the compressed latent domain, rather than directly on the full FE mesh representation. The tensor $H_t^3$ combines the encoded information of the current mesh state, provided by $Z_t$, with recurrent information accumulated from previous timesteps.

### 3.3.3 NED: The Node-Element Decoder Module

The final hidden state $H_t^3$ of the LTE module is decoded to produce the full mesh NEO using the NED module. Since NEOs are defined on different parts of the FE mesh, NED uses two dedicated prediction branches, *i.e.*, the node branch and the element branch (Fig. 5). First, the tensor $H_t^3$ is reshaped and is projected to the element domain by using adaptive average pooling. This results in an element-based tensor $H_t^{el} \in \mathbb{R}^{B \times F_h \times E}$, where $F_h$ denotes the feature channels, that is utilized by both branches of the decoder. The decoder is first tasked to predict the node output using the node branch. In this branch, the tensor $H_t^{el}$ is propagated through an element-to-node feature aggregation layer that averages the latent features of all elements that are incident to each node to produce the node-based latent tensor $H_t^n \in \mathbb{R}^{B \times F_h \times \Lambda}$ based on the FE mesh connectivity matrix $M$. Before the $H_t^n$ tensor is propagated through the node branch it is subsequently concatenated with the tensor $I_F^n$, to preserve sparse simulation-specific information that may be attenuated during encoding. The tensor $I_F^n$ is derived from the boundary and loading features in the input tensor $X_t$, Specifically, the mesh connectivity matrix $M$ is used to accumulate the loading and boundary features $(S_t, \Gamma)$ at the corresponding mesh nodes, where contributions assigned to the same node are summed to obtain $I_F^n \in \mathbb{R}^{B \times F_i \times \Lambda}$, where $F_i$ are the summed input features of $S_t$ and $\Gamma$.

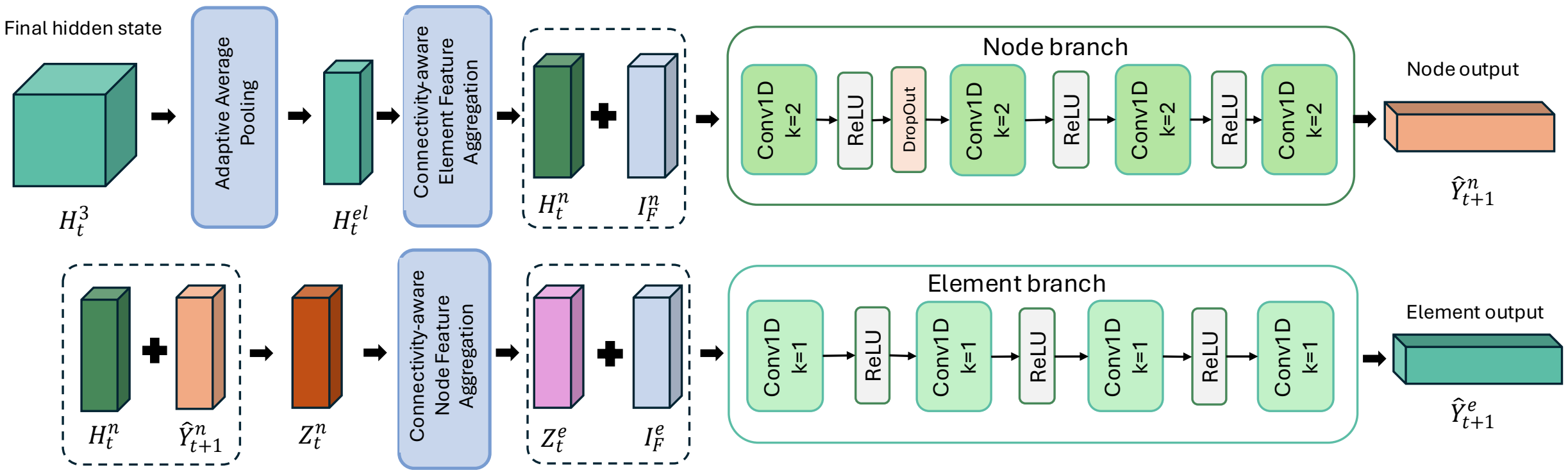

*Figure 5 Illustration of the NED dual-branch architecture.*

Then, a sequence of 1D convolutional layers predicts the node output $\hat{Y}_{t+1}^n \in \mathbb{R}^{B \times 3 \times \Lambda}$, which is the nodal displacement update. The updated predicted coordinates $\hat{N}_{t+1}$ for timestep $t+1$ are then obtained as follows:

$$\hat{N}_{t+1} = N_t + \hat{Y}_{t+1}^n \quad (4)$$

Subsequently, the output tensor $\hat{Y}_{t+1}^n$ is also utilized to predict the element output. The tensor $\hat{Y}_{t+1}^n$ is first concatenated with $H_t^n$ along the feature-channel dimension. The combined tensor $Z_{t+1}^n = \left[H_t^n, \hat{Y}_{t+1}^n\right] \in \mathbb{R}^{B \times (F_h+3) \times \Lambda}$ is then mapped to the element-based tensor $Z_{t+1}^e \in \mathbb{R}^{B_s \times (F_h+3) \times E}$ through a node-to-element feature aggregation layer using $M$. For each element $e$, the feature vectors of each node incident to $e$ are averaged as follows:

$$Z_{t+1}^e(b,:,e) = \frac{1}{G_e} \sum_{g=1}^{G_{\max}} V[e,g] \cdot Z_{t+1}^n(b,:,M[e,g]) \quad (5)$$

where $b$ indexes the batch sample, $G_e$ is the total number of nodes for element $e$, $g$ defines the local index of each node in $M$ for element $e$, and $M[e,g]$ returns the global mesh node index of the $g$-$th$ local node of element $e$, which is used to retrieve the corresponding feature vector from $Z_{t+1}^{n}$.

This yields a deformation-aware element tensor $Z_{t+1}^{e}$ that combines the latent temporal state with the predicted kinematic response. Then, $Z_{t+1}^{e}$ is further concatenated with the element-based input feature tensor $I_F^e \in \mathbb{R}^{B \times F_i \times E}$ that is formed by averaging $S_{\text{local}}^{t}$ and $\Gamma_{\text{local}}$ along the node group dimension $G_{\text{max}}$. During this aggregation, the validity mask $V$ is used to exclude padded entries from the aggregation process, such that each element is averaged only over its $G_e$ physical nodes. The resulting tensor is finally processed by a dedicated 1D convolutional decoder to produce $Y_{t+1}^{e}$, where the two output channels correspond to the effective stress and strain.

## 3.4 Training Procedure

The TopoNEP network is trained using datasets generated from conventional transient FEA simulations. Each simulation contains the time history of nodal coordinates and displacements, element-based effective stress and strain, boundary conditions, mesh connectivity, and simulation-specific input features such as forces, pressure, or other loading descriptors. During training, the network is tasked with predicting the output of the next timestep based on the current mesh state and the corresponding simulation-specific inputs. To predict the next timestep, the network relies on two distinct categories of input. The first category consists of external simulation parameters, such as boundary conditions, mesh connectivity, and loading descriptors (e.g., forces or pressure). These parameters are consistently provided to the network exactly as defined in the dataset. The second category is the evolving physical state of the mesh, specifically the nodal coordinates.

In this study, the training process expands upon the NELO algorithm introduced in (Triantafyllou et al., 2025). NELO is specifically designed to address the error accumulation problem that occurs in recursive surrogate models of transient FEA simulations. In detail, NELO controls how the nodal coordinates are fed into the network during training. At the beginning, the network receives the exact, ground truth nodal coordinates from the dataset. As training progresses, NELO gradually replaces these ground truth inputs with the network's own predicted coordinates from the preceding timestep. This gradual transition forces the model to adapt to its own prediction errors, successfully bridging the gap between training and the fully recursive conditions encountered during testing. The complete training algorithm can be observed in Algorithm 1.

**Algorithm 1: Training Procedure**

1: **Set** $P_s \leftarrow 1, k \leftarrow number\ of\ epochs\ to\ decrease\ P_s, \kappa \leftarrow 0, \gamma \leftarrow 0.7, \beta_s \leftarrow 0.01, S \leftarrow total\ epochs,$ $B \leftarrow total\ mini-batches$, and $T \leftarrow total\ timesteps.$
2: **for** epoch $j = 1$ to $S$ **do**
3:     **for** each batch $b = 1$ to $B$ **do**
4:         **for** timestep $t = 1$ to $T$ **do**
5:             Sample probability $P_r \sim Uniform\ (0, 1)$
6:             **if** $P_r > P_s$ and $t > 1$ **then**
7:                 **Set** $N_t = \widehat{N}_t$
8:             **end if**
9:             Set input tensor $X_t = (N_t, S_t, \Gamma)$
10:             Transform $X_t$ to $X_{\text{local}}^t$ using MCBG and ABEO
11:             $\hat{Y}_{t+1}^n, \hat{Y}_{t+1}^e = TopoNEP(X_{\text{local}}^t)$
                **Set** $\widehat{N}_t = N_t + \hat{Y}_{t+1}^n$
12:         **end for**
13:         Compute FEA-informed objective $L_{FNE}$ over all $T$ timesteps
14:         Update TopoNEP parameters $\theta$ using an optimization step
15:     **end for**
16:     **if** $j \ \%\ k == 0$ **then**
17:         $\kappa = \kappa + 1$
18:         $P_s = \gamma^\kappa$
19:     **end if**
20:     **if** $P_s < \beta_s$ **then**
21:         $P_s = 0$
22:     **end if**
23: **end for**

### 3.4.1 FEA-Informed Node-Element Loss Optimization

Considering the dual-branch topology of TopoNEP, the training objective must account for all NEO prediction errors. In the previous DeepFEA framework (Triantafyllou et al., 2025), this was achieved through the Node-Element loss, which jointly optimized the NEO prediction errors. In DeepFEAv2, the same principle is retained, but the loss function is extended with additional FEA-informed loss components that improve recursive stability, preserve local mesh geometry, and reduce non-physical deformation artifacts. The resulting $L_{\text{FNE}}$ loss function is defined as:

$$L_{\text{FNE}} = w_{\text{dis}} L_{\text{dis}} + \lambda_{\text{eff}} L_{\text{eff}} + w_{\text{edge}} L_{\text{edge}} + w_{\text{norm}} L_{\text{norm}} + w_{\text{time}} L_{\text{time}} + w_{\text{vel}} L_{\text{vel}} \quad (6)$$

where $L_{\text{dis}}$ is the nodal displacement loss, $L_{\text{eff}}$ is the effective stress and strain loss, $L_{\text{edge}}$ is the edge-length regularization loss, $L_{\text{norm}}$ is the surface-normal consistency loss, $L_{\text{time}}$ is the temporal velocity loss, and $L_{\text{vel}}$ is the local velocity-smoothing loss. The coefficients $w_{\text{dis}}$, $w_{\text{edge}}$, $w_{\text{norm}}$, $w_{\text{time}}$, and $w_{\text{vel}}$ control the relative contribution of each term. The coefficient $\lambda_{\text{eff}}$ is dynamically adjusted to balance the element-based loss with the node-based loss.

The $L_{\mathrm{FNE}}$ loss is computed over the entire predicted output sequence. Each loss component accumulates the corresponding prediction error over all predicted timesteps $T$ and over all the mesh entities on which that quantity is defined. Specifically, nodal displacement errors are evaluated over all nodes, effective stress and strain errors over all elements, edge-length errors over mesh edges, surface-normal errors over surface faces, and temporal consistency errors over consecutive predicted states. The resulting terms are averaged over their corresponding timesteps and mesh entities so that the total objective is not dominated by the number of nodes, elements, or predicted timesteps. In detail, the $L_{dis}$ loss is defined as:

$$L_{dis} = \frac{1}{T \cdot \Lambda} \sum_{t=1}^{T} \sum_{i=1}^{\Lambda} \omega_i^t \cdot \| \hat{u}_i^t - u_i^t \|_1 \tag{7}$$

where $u_i^t$ and $\hat{u}_i^t$ are the ground truth and predicted displacement vectors of node $i$ at timestep $t$, respectively. The weight $\omega_i^t$ increases the penalty for nodes undergoing larger motion and is defined as:

$$\omega_i^t = 1 + \alpha_\omega \| u_i^t - u_i^{t-1} \|_2 \tag{8}$$

where $\alpha_\omega$ is a scalar weighting coefficient, $\|\cdot\|_2$ denotes the Euclidean norm and $\omega_i^t$ is the displacement-loss weight assigned to node $i$ at timestep $t$. Thus, $L_{\mathrm{dis}}$ encourages accurate nodal deformation prediction, with greater emphasis on highly dynamic regions. The $L_{\mathrm{eff}}$ loss is defined as:

$$L_{\mathrm{eff}} = \frac{1}{T \cdot E} \sum_{t=1}^{T} \sum_{e=1}^{E} \| \hat{Y}_e^t - Y_e^t \|_2^2 \tag{9}$$

where $Y_e^t = (\Sigma_e^t, E_e^t)$ contains the ground truth effective stress and strain of element $e$ and $\hat{Y}_e^t$ is the corresponding prediction. This term trains the element branch to recover the material response of the simulation. Since the magnitude of stress and strain errors can differ substantially from that of displacement errors, the element loss is adaptively scaled as:

$$\lambda_{\mathrm{eff}} = \mathrm{clamp}\left(\frac{\beta \cdot L_{\mathrm{dis}}}{L_{\mathrm{eff}} + \epsilon_{eff}}, \lambda_{\mathrm{min}}, \lambda_{\mathrm{max}}\right) \tag{10}$$

where $\beta$ is a scalar scaling coefficient, $\epsilon_{\mathrm{eff}}$ is a small positive numerical constant, and $\lambda_{\mathrm{min}}$ and $\lambda_{\mathrm{max}}$ are scalar lower and upper bounds for $\lambda_{\mathrm{eff}}$, respectively. This follows the node-element balancing principle of NELO. The edge-length loss is defined as:

$$L_{\mathrm{edge}} = \frac{1}{T \cdot | \mathcal{E}_{\mathrm{edge-5}} |} \sum_{t=1}^{T} \sum_{(i,j)\in\mathcal{E}_{\mathrm{edge-5}}} \left( \frac{\| \hat{x}_i^t - \hat{x}_j^t \|_2 - \| x_i^t - x_j^t \|_2}{\| x_i^t - x_j^t \|_2 + \epsilon_{edge}} \right) \tag{11}$$

where $x_i^t$ and $\hat{x}_i^t$ denote the ground truth and predicted coordinates of node $i$, respectively. The set $\mathcal{E}_{\text{edge-5}}$ contains the top 5% most distorted edges of the FE mesh, and $\epsilon_{\mathrm{edge}}$ is a small positive constant that prevents numerical instability when normalizing by very small edge lengths. This term penalizes non-physical stretching or compression of connected nodes. The $L_{\mathrm{norm}}$ loss is utilized only for meshes composed of shell elements and is defined as:

$$L_{\text{norm}} = \frac{1}{T \cdot |\mathcal{F}_{10}|} \sum_{t=1}^{T} \sum_{k \in \mathcal{F}_{10}} \left(1 - \frac{\hat{n}_k^t \cdot n_k^t}{\|\hat{n}_k^t\|_2 \|n_k^t\|_2 + \epsilon_{\text{norm}}}\right) \tag{12}$$

where $n_k^t$ and $\hat{n}_k^t$ are the ground truth and predicted normal vectors of face $k$ at timestep $t$, respectively and $\|\cdot\|_2$ denote the Euclidean norm of each vector. The set $\mathcal{F}_{10}$ contains the top 10% surface faces with the largest normal inconsistencies and $\epsilon_{\text{norm}}$ is a small positive numerical constant. In addition, $|\mathcal{F}_{10}|$ denotes the number of faces in $\mathcal{F}_{10}$. Overall, $L_{\text{norm}}$ penalizes changes in surface orientation and helps reduce folding or inversion artifacts. For simulations exhibiting strong or rapidly varying damping, the temporal velocity and velocity-smoothing losses are included to better constrain the predicted transient response. The temporal velocity loss is defined as:

$$L_{\text{time}} = \frac{1}{(T-1) \cdot \Lambda} \sum_{t=2}^{T} \sum_{i=1}^{\Lambda} \|\hat{v}_i^t - v_i^t\|_1 \tag{13}$$

where $\hat{v}_i^t = \hat{x}_i^t - \hat{x}_i^{t-1}$ and $v_i^t = x_i^t - x_i^{t-1}$. This term encourages the predicted trajectory to follow the correct temporal evolution. Finally, the velocity-smoothing loss is defined as:

$$L_{\text{vel}} = \frac{1}{(T-1) \cdot |\mathcal{E}_{\text{edge}}|} \sum_{t=2}^{T} \sum_{(i,j) \in \mathcal{E}_{\text{edge}}} \|\hat{v}_i^t - \hat{v}_j^t\|_2^2 \tag{14}$$

where $\hat{v}_i^t$ and $\hat{v}_j^t$ are the predicted finite-difference velocity vectors of adjacent nodes $i$ and $j$, respectively, and $(i,j) \in \mathcal{E}_{\text{edge}}$ denotes an edge connecting the two nodes in the FE mesh. This term suppresses nonphysical high-frequency oscillations between neighboring nodes during recursive prediction. Overall, the proposed objective extends the previous DeepFEA node-element loss by combining NEO prediction accuracy, mesh-shape preservation, and temporal consistency in a single optimization framework.

# 4 Experiments and Results

To systematically evaluate the predictive accuracy, applicability across different mesh topologies, and computational efficiency of DeepFEAv2, a comprehensive series of experiments was performed across varying mesh topologies, element formulations, and loading conditions. First, DeepFEAv2 was evaluated on two 3D Linear Elastic Model (LEM) datasets, with structured and unstructured meshes, to systematically assess model performance under controlled conditions while increasing topological complexity. Second, to demonstrate the model's capacity to generalize to realistic engineering and biomechanical settings, an application study was conducted on a dynamic, pressure-driven aortic valve simulation featuring hyperelastic material behavior and shell elements. Finally, the inference speedups achieved by DeepFEAv2 were evaluated against conventional FEA solvers.

## 4.1 Evaluation Metrics

The performance of DeepFEAv2 was evaluated by measuring the accuracy of both NEO predictions. The coefficient of determination $R^2$, normalized Mean Absolute Error (NMAE), and normalized Root Mean Squared Error (NRMSE) were used as the main metrics for quantitative evaluation. The $R^2$ metric, known as coefficient of determination, has been selected, since it is widely used for the evaluation of regression models (Leach and Henson, 2007). In the context of FEA surrogate modeling, $R^2$ provides a statistical measure of how well the predicted NEO aligns with the corresponding ground truth obtained from FEA simulations as described in Subsection 4.2. The coefficient of determination is defined as:

$$R^2 = 1 - \frac{\sum_j (y_j - \hat{y}_j)^2}{\sum_j (y_j - \bar{y})^2} \tag{15}$$

where $y_j$ is the $j$-th ground truth, $\hat{y}_j$ is the corresponding prediction, and $\bar{y}$ is the mean of the ground truth. An $R^2$ value close to 1 indicates that the surrogate predictions closely match the FEA ground truth data, whereas an $R^2$ value close to 0 corresponds to a baseline model that predicts the mean of the ground truth values. Negative $R^2$ values may occur when the model performs worse than this baseline.

The evaluation of FEA response fields presents an additional challenge because different physical quantities may have substantially different magnitudes and units. For example, nodal displacement, effective stress, and effective strain can vary over different numerical ranges, while stress values may span several orders of magnitude depending on the material model and loading conditions. In such cases, non-normalized error metrics, such as MAE or RMSE, are difficult to compare directly because their values depend on the physical scale and unit of each output quantity. Therefore, NMAE and NRMSE were also included in the evaluation to express the prediction error relative to the range of the corresponding ground truth. These normalized metrics provide scale-independent error measures and have been reported to be more robust than their non-normalized counterparts when assessing predictors in cases where output variables have different magnitudes (Liang et al., 2020). The NMAE for a predicted physical parameter $p$ is defined as:

$$\mathrm{NMAE}_p = \frac{1}{A}\sum_{j=1}^{A} \frac{\sum_{\mu=1}^{\Phi} | y_{\mu j} - \hat{y}_{\mu j} |}{\Phi \left(\max\{Y_j^p\} - \min\{Y_j^p\}\right)} \times 100\% \tag{16}$$

where $A$ is the total number of simulations, $\Phi$ is the number of output values for simulation $j$, and $Y_j^p$ denotes the set of ground truth values of parameter $p$ for simulation $j$. Physically, NMAE represents the average absolute prediction error as a percentage of the response range of the corresponding physical quantity. The NRMSE is defined as:

$$\mathrm{NRMSE}_p = \frac{1}{A}\sum_{j=1}^{A} \frac{\sqrt{\sum_{\mu=1}^{\Phi} (y_{\mu j} - \hat{y}_{\mu j})^2}}{\sqrt{\Phi}\left(\max\{Y_j^p\} - \min\{Y_j^p\}\right)} \times 100\% \tag{17}$$

Physically, NRMSE represents the root mean squared prediction error as a percentage of the response range. Since squared errors are used, NRMSE is more sensitive than NMAE to larger local deviations in the predicted displacement, stress, or strain fields.

## 4.2 Transient FEA for Structured and Unstructured Meshes

DeepFEAv2 is a generalized framework that can be applied to transient FEA problems involving structured or unstructured meshes. For the main experimental evaluation, two 3D linear elastic material datasets were used. These datasets were selected to assess DeepFEAv2 under controlled conditions while progressively increasing the complexity of mesh representation. The first dataset is a Structured 3D LEM that uses a regular volumetric mesh and therefore provides a direct comparison point with the previous DeepFEA framework (Fig. 6a). The second dataset, referred to as Unstructured 3D LEM, follows the same general solid-mechanics prediction task but replaces the regular element arrangement with a substantially denser unstructured mesh (Fig. 6b). Together, these two datasets allow the proposed model to be evaluated on both structured and unstructured FE representations.

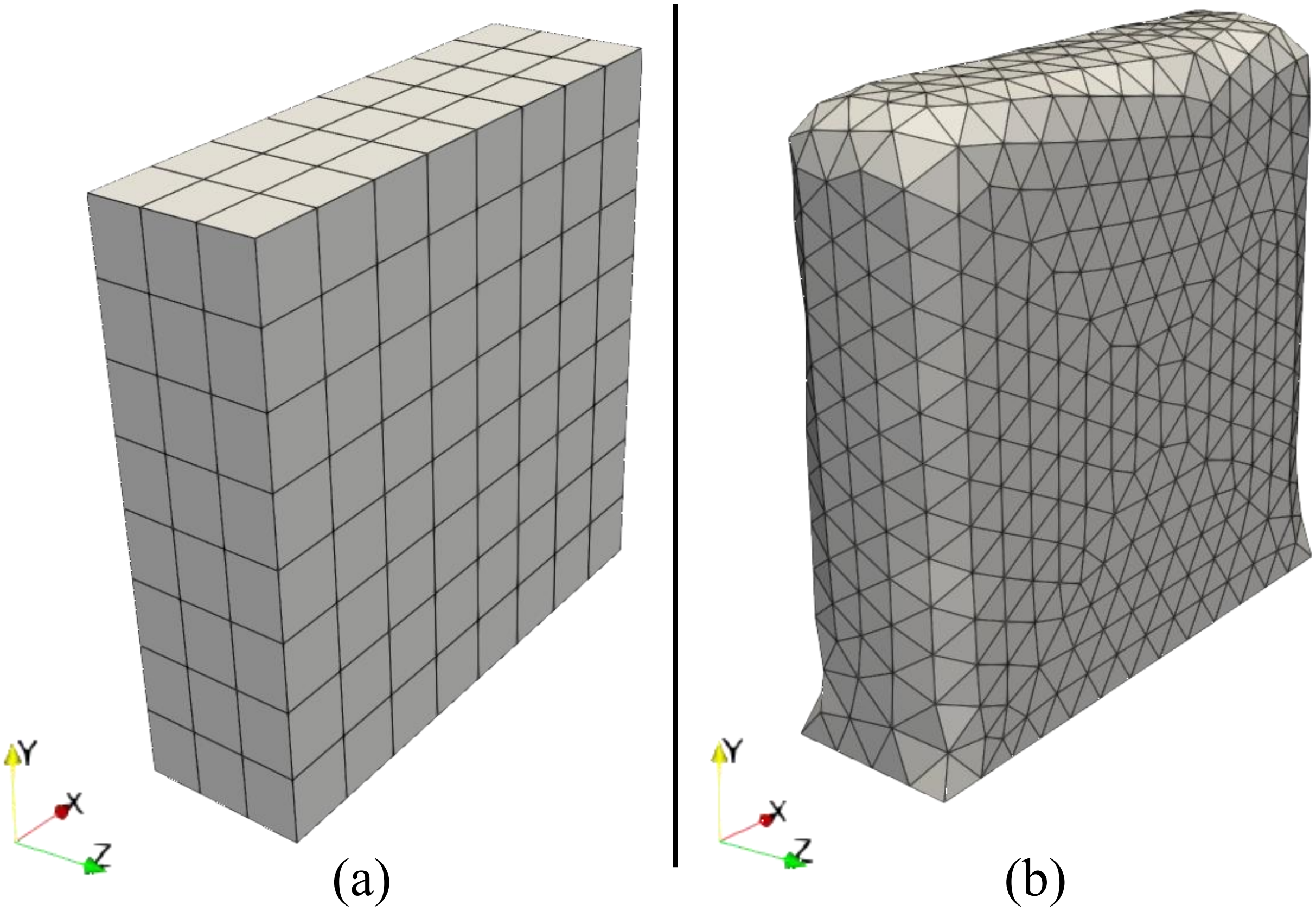

*Figure 6 Illustrations of the FE meshes in (a) the 3D structured and (b) 3D unstructured LEM datasets.*

Both datasets were generated using transient simulations produced by LS-DYNA R15.0.2, a commercial software provided by ANSYS. At each selected timestep, the model receives the current nodal geometry, boundary-condition information, FE connectivity, and force-related input features. The supervised outputs are the updated nodal coordinates or displacements and the element-based effective stress and strain. The effective stress and strain were computed from the stress and strain tensor components using von Mises-type formulations. The effective stress is defined as:

$$\sigma_{\text{eff}} = \sqrt{\frac{(\sigma_{xx} - \sigma_{yy})^2 + (\sigma_{yy} - \sigma_{zz})^2 + (\sigma_{zz} - \sigma_{xx})^2 + 6(\sigma_{xy}^2 + \sigma_{yz}^2 + \sigma_{zx}^2)}{2}} \tag{18}$$

where $\sigma_{\text{eff}}$ denotes the effective stress and $\sigma_{xx}, \sigma_{yy}, \sigma_{zz}, \sigma_{xy}, \sigma_{yz}, \sigma_{zx}$ are the stress tensor components.

The effective strain is defined as:

$$\varepsilon_{\text{eff}} = \frac{2}{3\sqrt{2}}\sqrt{(\varepsilon_{xx} - \varepsilon_{yy})^2 + (\varepsilon_{yy} - \varepsilon_{zz})^2 + (\varepsilon_{zz} - \varepsilon_{xx})^2 + 6(\varepsilon_{xy}^2 + \varepsilon_{yz}^2 + \varepsilon_{zx}^2)} \tag{19}$$

where $\varepsilon_{\text{eff}}$ denotes the effective strain and $\varepsilon_{xx}, \varepsilon_{yy}, \varepsilon_{zz}, \varepsilon_{xy}, \varepsilon_{yz}, \varepsilon_{zx}$ are the strain tensor components.

The Structured 3D LEM dataset was used as the controlled baseline because it preserves the regular mesh organization of the previous DeepFEA. The mesh comprises a $9 \times 9 \times 4$ grid, corresponding to 324 nodes and 192 eight-node (hexahedral) solid elements. The material is linear elastic, with a density of $1200\,\text{kg/m}^3$, Young's modulus of $5 \times 10^6$ Pa, and Poisson's ratio of 0.495. The dataset consists of force-driven simulations where external nodal forces are applied to selected unconstrained surface nodes, while the bottom boundary nodes remain fixed. A set of 480 simulations was used, with each simulation comprising 30 timesteps, resulting in 14,400 spatiotemporal samples. The input channels include dynamic nodal coordinates, sparse force components, force-context descriptors, and a flag that denotes bounded or free-nodes.

The Unstructured 3D LEM dataset extends the same force-driven prediction task to a larger unstructured mesh. It contains 962 nodes and 3,669 four-node (tetrahedral) solid elements. The dataset includes 80 simulations with 30 timesteps each, resulting in 2,400 spatiotemporal samples (80 $simulations \cdot 30\ timesteps$). The loading conditions, output quantities, and input features follow the structured dataset, including sparse force channels, dense force-context descriptors, and the bounded/free-node flag.

### 4.2.1 Experimental Setup

The performance of DeepFEAv2 was evaluated in the context of predicting nodal displacements, effective stress, and effective strain. Each dataset was first split into training and testing subsets using an 80%/20% ratio at the simulation level. The training subset was further divided into internal training and validation subsets using an 80%/20% ratio. The internal validation subset was used for hyperparameter selection and best-checkpoint selection. The proposed framework was implemented in PyTorch and trained for 500 epochs with a batch size of 16 for the structured 3D LEM and 8 for the unstructured 3D LEM, due to the larger mesh size and GPU memory constraints. The LTE module operated an 8×8×8 latent spatial grid. Optimization was performed using AdamW with a learning rate of $3\times10^{-4}$, weight decay of $1\times10^{-4}$, and a sequential learning rate schedule consisting of a 10% warmup phase with constant learning rate followed by cosine annealing decay to a minimum of $1\times10^{-6}$ over the remaining epochs. The loss weights were defined as $w_{\text{dis}} = 10$, $w_{\text{edge}} = 5$, $w_{\text{time}} = 2$, and $w_{\text{vel}} = 1$, as these yielded the best overall prediction accuracy. The $L_{\text{norm}}$ was disabled by setting $w_{\text{norm}} = 0$, since these datasets use solid elements rather than shell elements. In addition, loss-specific parameters were defined as $\beta = 10$, $\epsilon_{\text{eff}} = 10^{-8}$, $\lambda_{\min} = 0.1$, $\lambda_{\max} = 10^3$, $\alpha_\omega = 15.0$, $\epsilon_{\text{edge}} = 10^{-8}$ and $\epsilon_{\text{norm}} = 10^{-8}$. DeepFEAv2 was also compared with the previous DeepFEA framework on the Structured 3D LEM dataset. This comparison was limited to the structured benchmark because the previous DeepFEA relies on regular tensor representations and cannot be directly applied to the unstructured LEM dataset. Therefore, the Structured 3D LEM dataset served as the common reference case for evaluating whether DeepFEAv2 preserves or improves the predictive performance of the previous framework while extending its applicability to general mesh topologies.

### 4.2.2 Structured 3D LEM

The quantitative results on the Structured 3D LEM dataset demonstrate that DeepFEAv2 improves the predictive performance of the previous DeepFEA framework while introducing the ability to operate through MCBG. As shown in Table 1, DeepFEAv2 achieves higher accuracy across all evaluated quantities, including nodal displacement and the element-wise effective stress and effective strain fields. For nodal displacement prediction, DeepFEAv2 substantially improves the coefficient of determination, increasing the displacement $R^2$ from 0.71 for DeepFEA to 0.98, corresponding to a relative increase of 38.0%. This improvement is also reflected in the normalized error metrics, where the displacement NMAE decreases from 3.40% to 0.44%, corresponding to an 87.1% relative error reduction, and the NRMSE decreases from 4.73% to 0.91%, corresponding to an 80.8% relative error reduction. These results indicate that the proposed topology-aware formulation provides a more accurate prediction of the recurrent kinematic variable. This is particularly important in transient FEA simulation settings, where displacement predictions are recursively used to estimate the system state at subsequent time steps.

DeepFEAv2 also achieves a clear improvement for the element-level mechanical quantities. For effective stress, $R^2$ increases from 0.75 with DeepFEA to 0.97, corresponding to a 29.3% relative increase. The effective stress NMAE decreases from 1.80% to 0.78%, corresponding to a 56.7% relative error reduction, while the NRMSE decreases from 2.96% to 1.34%, corresponding to a 54.7% relative error reduction. A similar improvement is observed for effective strain, where $R^2$ increases from 0.75 to 0.97, corresponding to a 29.3% relative increase. The effective strain NMAE decreases from 1.79% to 0.78%, corresponding to a 56.4% relative error reduction, and the NRMSE decreases from 2.95% to 1.34%, corresponding to a 54.6% relative error reduction. These results show that DeepFEAv2 not only captures the global variance of the element-level response more effectively but also reduces the pointwise prediction errors for stress and strain. For linear elastic models such as the one used in the Structured 3D LEM dataset, effective stress and effective strain are proportional. Notably, DeepFEA and DeepFEAv2 preserve this proportionality in their predictions, leading to nearly identical NMAE, NRMSE, and $R^2$ values for both effective stress and strain outputs.

*Table 1 Quantitative results for the structured 3D LEM dataset, comparing DeepFEA with DeepFEAv2.*

| Methods | Displacement | | | Effective Stress | | | Effective Strain | | |
|---|---|---|---|---|---|---|---|---|---|
| | $R^2$ ↑ | NMAE (%) ↓ | NRMSE (%) ↓ | $R^2$ ↑ | NMAE (%)↓ | NRMSE (%) ↓ | $R^2$ ↑ | NMAE (%) ↓ | NRMSE (%) ↓ |
| DeepFEA | 0.71 | 3.40 | 4.73 | 0.75 | 1.80 | 2.96 | 0.75 | 1.79 | 2.95 |
| DeepFEAv2 | 0.98 | 0.44 | 0.91 | 0.97 | 0.78 | 1.34 | 0.97 | 0.78 | 1.34 |

Note: ↑ indicates better performance for larger values and ↓ for smaller ones.

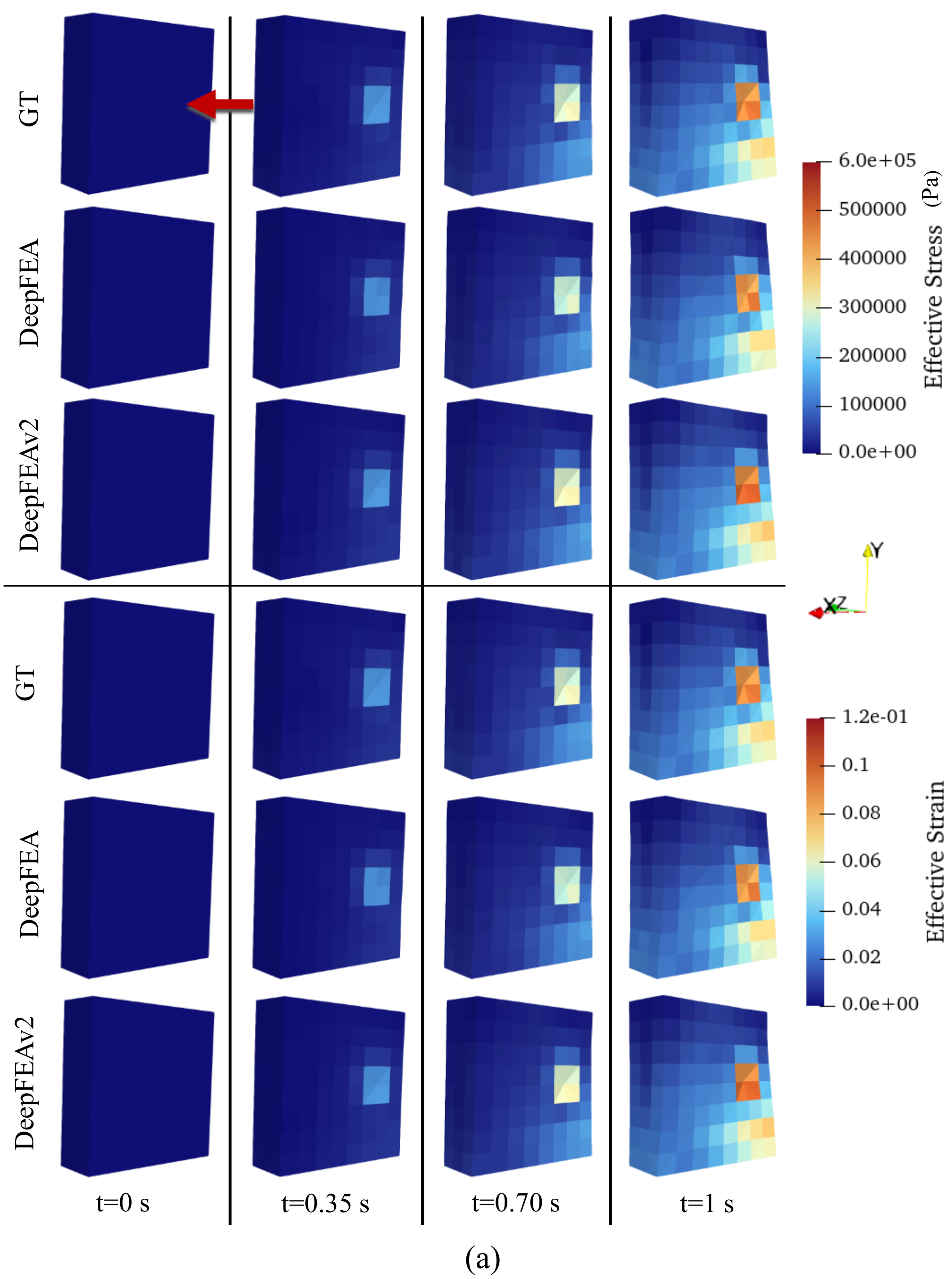
GT
DeepFEA
DeepFEAv2
GT
DeepFEA
DeepFEAv2
t=0 s
t=0.35 s
t=0.70 s
t=1 s
6.0e+05
500000
400000
300000
200000
100000
0.0e+00
Effective Stress (Pa)
1.2e-01
0.1
0.08
0.06
0.04
0.02
0.0e+00
Effective Strain
Y
X
Z

(a)

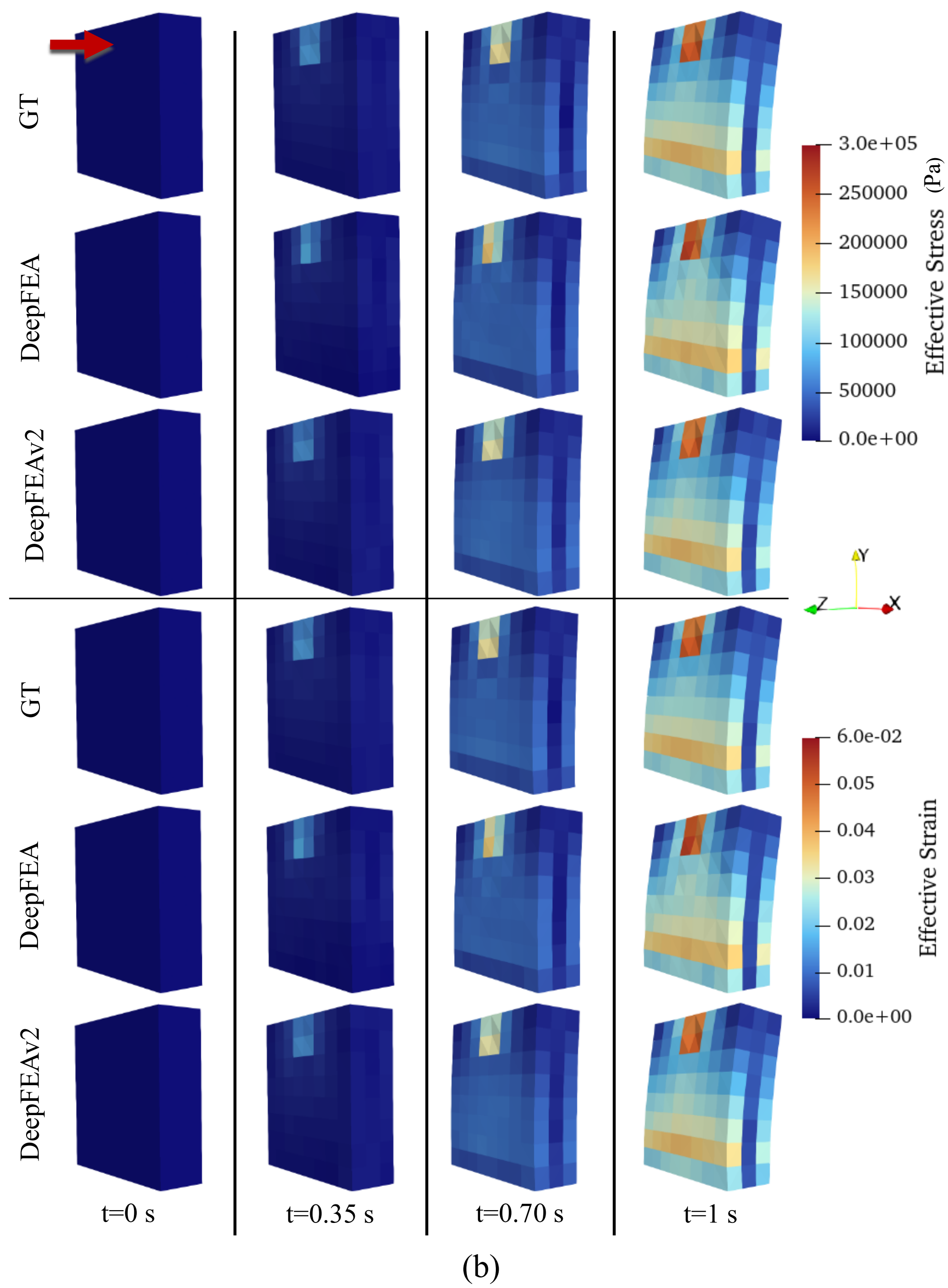

*Figure 7 Qualitative results for two unknown simulations (a) and (b), depicting effective stress in Pa and strain, where GT denotes the ground truth and the red arrows indicate the applied load.*

The qualitative results in Fig. 7 show two unknown simulation cases from the Structured 3D LEM dataset. For each simulation, the ground truth, DeepFEA, and the proposed DeepFEAv2 model are compared over multiple time steps. In each visualization, the deformed shape shows the predicted or simulated displacement of the structure over time, while the color maps represent the magnitude of the evaluated mechanical field. Cooler colors, such as dark blue, correspond to lower effective stress or strain values, whereas warmer colors, such as yellow, orange, and red, indicate regions with higher mechanical response. The apparent tilting or bending of the block is caused by the applied nodal force acting on one side of the structure while the bottom boundary remains constrained, producing a progressive deformation pattern during the transient simulation.

The visual comparison demonstrates that both models capture the general evolution of the structure under loading *i.e.*, the externally applied nodal force indicated by the red arrows. Nevertheless, DeepFEAv2 provides a closer agreement with the ground truth, particularly in the regions where the mechanical response becomes more localized and intense. In Fig. 7a, both DeepFEA and DeepFEAv2 reproduce the overall temporal progression of the response from the initial low-stress and low-strain state to the later loading stages. During the temporal simulation progression, a localized high-response region develops near the force application area. DeepFEA captures the main deformation trend and the approximate location of the high-response region. However, DeepFEAv2 provides a more accurate reconstruction of the effective stress and effective strain distributions. In particular, the predicted high-response region and the surrounding gradient are more consistent with the ground truth, while the broader low- and medium-intensity regions are also better preserved. This suggests that the MCBG improves the prediction of the mechanical field distribution, even in cases where the structured-grid DeepFEA model already performs reasonably well. The second representative case highlights the improvement more clearly, Fig. 7b. In this simulation, DeepFEA shows larger discrepancies in the loaded regions, where the deformation is larger. The predicted motion and the corresponding effective stress and strain fields appear less consistent with the ground truth, especially around the localized high-response areas (near the upper loaded region) and the broader regions of elevated mechanical response that develop as the simulation progresses. By contrast, DeepFEAv2 better captures the location, intensity, and spatial extent of the loaded regions. As a result, the effective stress and strain fields predicted by DeepFEAv2 follow the ground truth patterns more closely across the evaluated time steps.

Overall, the qualitative comparison shows that DeepFEAv2 improves upon DeepFEA in both representative simulations. While DeepFEA can reproduce the general response pattern, its predictions are less accurate in regions with stronger deformation or higher stress and strain concentration. DeepFEAv2 produces more consistent displacement-driven motion and more accurate element-wise effective outputs, particularly in the loaded regions where local mechanical gradients are more difficult to predict. These visual observations support the quantitative results in Table 1 and indicate that the proposed MCBG formulation improves the modeling of transient mechanical behavior in structured 3D FEA simulations.

### 4.2.3 Unstructured 3D LEM

Table 2 presents the quantitative results of DeepFEAv2 on the Unstructured 3D LEM dataset. The model achieved an $R^2$ of 0.97 for displacement prediction, with an NMAE of 1.42% and an NRMSE of 2.57%. For the element-level outputs, DeepFEAv2 obtained an $R^2$ of 0.84 for both effective stress and strain. The corresponding NMAE and NRMSE values were 1.55% and 3.06% for both quantities. As in the structured LEM dataset, the Unstructured 3D LEM dataset uses the same linear elastic material model and material properties. Effective stress and effective strain are therefore proportional, and DeepFEAv2 preserves this proportionality in its predictions, leading to nearly identical NMAE, NRMSE, and $R^2$ values. Overall, these results show that DeepFEAv2 can predict the transient response of a larger unstructured mesh using explicit MCBG, without requiring a regular grid-based representation.

*Table 2 Quantitative results for the unstructured 3D LEM dataset.*

| Methods | Displacement | | | Effective Stress | | | Effective Strain | | |
|---|---|---|---|---|---|---|---|---|---|
| | $R^2$ ↑ | NMAE (%) ↓ | NRMSE (%) ↓ | $R^2$ ↑ | NMAE (%) ↓ | NRMSE (%) ↓ | $R^2$ ↑ | NMAE (%) ↓ | NRMSE (%) ↓ |
| DeepFEAv2 | 0.97 | 1.42 | 2.57 | 0.84 | 1.55 | 3.06 | 0.84 | 1.55 | 3.06 |

Note: ↑ indicates better performance for larger values and ↓ for smaller ones.

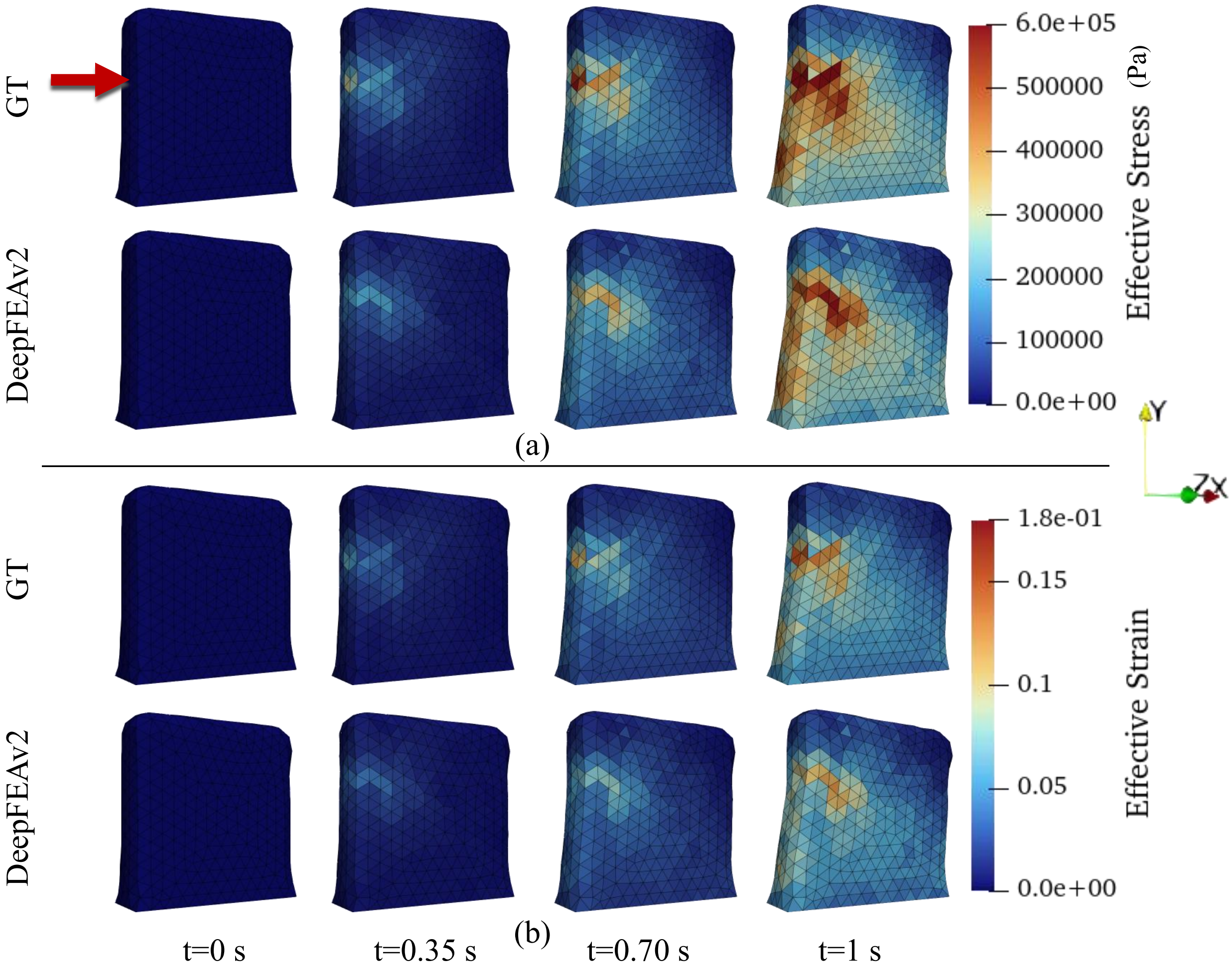

*Figure 8 Qualitative results for a representative simulation of the unstructured dataset, depicting (a) effective stress in Pa and (b) effective strain, where GT denotes the ground truth and the red arrow indicates the applied load.*

The qualitative results in Fig. 8 show the progression of one unknown simulation from the Unstructured 3D LEM dataset at different time steps. In this visualization, the red arrow indicates the externally applied nodal force, while the deformation of the object shows the displacement-driven motion of the unstructured mesh over time. The color maps represent the magnitude of effective stress or effective strain, with warmer colors indicating higher mechanical response values. The predicted effective stress and effective strain fields closely follow the global spatial patterns observed in the ground truth. Specifically, DeepFEAv2 captures the progressive increase in mechanical response over time and reproduces the main high-response regions near the force application area and constrained regions of the structure. Since the dataset is generated from a linear elastic formulation, the stress and strain fields evolve smoothly and remain spatially consistent with the displacement-driven deformation pattern. Although small local discrepancies are visible in regions with higher response intensity close to the externally applied force, the predicted fields preserve the overall distribution, temporal progression, and relative concentration of stress and strain across the unstructured mesh. These qualitative observations are consistent with the quantitative results and demonstrate that the proposed MCBG formulation can model transient behavior on non-grid-compatible FE meshes.

## 4.3 Application Study

To further assess the applicability of DeepFEAv2 beyond controlled benchmark cases, an additional application study was performed on a pressure-driven aortic valve dataset. The aortic valve is a critical cardiac structure that regulates blood flow from the left ventricle into the aorta and undergoes large transient deformation under time-varying pressure loads. Accurate prediction of its deformation, stress, and strain fields is important for understanding valve biomechanics and for supporting the design and assessment of surgical or prosthetic interventions. This dataset was selected since it differs substantially from the LEM datasets in geometry, element type, loading mechanism, and physical complexity. In particular the aortic valve dataset consists of transient nonlinear shell-element-based FEA simulations of a pressure-driven hyperelastic aortic valve model. Therefore, it provides a realistic test case for evaluating whether DeepFEAv2 can generalize beyond structured and unstructured solid linear elasticity simulations to more complex shell-based biomechanical simulations.

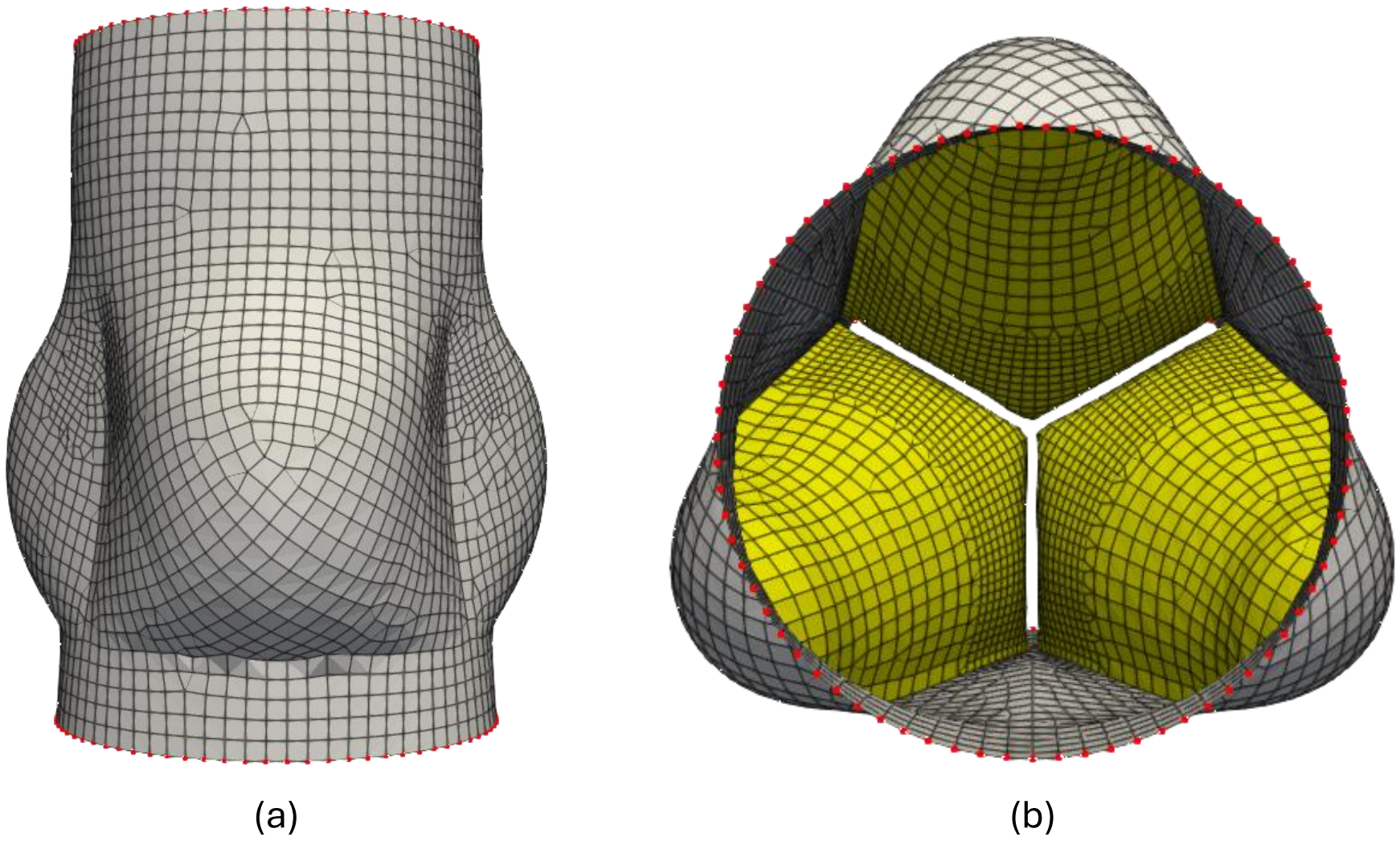

*Figure 9 Illustration of the aortic valve FEM used in the pressure-driven application study. The left view (a) shows the external side view of the valve model, while the right view (b) shows the top view of the leaflet configuration and root boundary. The grey*

*surface represents the valve root, the yellow surfaces represent the three valve leaflets, and the red markers indicate the constrained boundary region of the root.*

The aortic valve simulations involved shell elements and pressure loading rather than solid elements and sparse nodal point forces. The aortic valve mesh comprised 5,533 nodes and 5,649 three- and four-node shell elements (Kalozoumis, 2018). The dataset contained 20 simulations that differed in the peak magnitude of the applied ventricular pressure load, which ranged from $5.0 \times 10^3$ to $1.6 \times 10^4$ Pa, while the geometry, mesh, material definitions, and boundary conditions were kept fixed. The valve leaflets were modeled using a soft-tissue hyperelastic material with density $\rho = 1{,}200\ \mathrm{kg/m^3}$, $C_1 = 6.0 \times 10^4$ Pa, $C_2 = 0$, $C_3 = 1.174 \times 10^5$ Pa, $C_4 = 9.19$, and bulk modulus $K = 3.0 \times 10^8$ Pa. The leaflet shell thickness was $5.0 \times 10^{-4}$ m. The root region was modeled as a nearly incompressible Mooney-Rivlin hyperelastic material with $\rho = 1{,}200\ \mathrm{kg/m^3}$, Poisson's ratio $\nu = 0.498$, coefficient $A = 1.0 \times 10^5$ Pa, $B = 0$, and shell thickness $2.0 \times 10^{-3}$ m. The aortic valve simulation outputs were nodal displacements and element-based effective stress and strain. Shell stress and strain components were averaged over each shell's integration points before computing the effective quantities. Subsequently, the effective stress and strain arrays were reordered according to the same element order used for the model input. The training process followed the same configuration as the unstructured 3D LEM experiments with one exception. Unlike the solid-element LEM experiments, the surface-normal consistency loss $L_{\mathrm{norm}}$ was also included to preserve local surface orientation deformation, which is important for shell-based meshes. As such, the $w_{\mathrm{norm}}$ was set to 10, since it provided the best validation performance. Since the aortic valve consists of shell elements, adjacency-based ordering was performed using shared edges rather than shared solid faces. This produces a topologically coherent element sequence suitable for the 1D convolutional encoder, even though the physical geometry is curved, thin, and composed of multiple anatomical parts including the valve root and three leaflets, as illustrated in Fig. 9.

The pressure-driven aortic valve application study provides the strongest preliminary evidence that DeepFEAv2 can generalize beyond the LEM datasets. Despite the increased geometric and physical complexity of the aortic valve dataset, DeepFEAv2 achieved high predictive accuracy across all output quantities (Table 3). The model obtained an $R^2$ of 0.99 for displacement, 0.98 for effective stress, and 0.98 for effective strain. The corresponding normalized errors were also low, with NMAE values of 0.38%, 0.65%, and 0.83% for displacement, effective stress, and effective strain, respectively. Similarly, the NRMSE values remained low across all outputs, reaching 1.18% for displacement, 1.37% for effective stress, and 1.51% for effective strain. Unlike the LEM datasets, the aortic valve simulations involve hyperelastic material behavior in both the leaflets and the root region. As a result, effective stress and effective strain are governed by nonlinear and region-dependent material relations rather than by a single constant scale factor. Consistent with this non-proportional behavior, DeepFEAv2 yields different NMAE and NRMSE values for the two outputs, rather than the nearly identical values observed for the LEM datasets. Overall, these results indicate that the proposed framework can accurately model transient deformation and element-level mechanical response in a pressure-driven shell structure.

*Table 3 Quantitative results for the aortic valve dataset.*

| Methods | Displacement | | | Effective Stress | | | Effective Strain | | |
|---|---|---|---|---|---|---|---|---|---|
| | $R^2$ ↑ | NMAE (%) ↓ | NRMSE (%) ↓ | $R^2$ ↑ | NMAE (%) ↓ | NRMSE (%) ↓ | $R^2$ ↑ | NMAE (%) ↓ | NRMSE (%) ↓ |
| DeepFEAv2 | 0.99 | 0.38 | 1.18 | 0.98 | 0.65 | 1.37 | 0.98 | 0.83 | 1.51 |

Note: ↑ indicates better performance for larger values and ↓ for smaller ones.

The qualitative results in Figs. 10 and 11 show representative predictions from the aortic valve application study. Figure 10 presents the leaflet-level results from both top and side views, allowing the predicted mechanical response to be evaluated from complementary viewing directions. DeepFEAv2 closely follows the ground truth deformation throughout the pressure-driven loading sequence, capturing the progressive opening, bending, and spatial evolution of the valve leaflets from the initial configuration to the later pressurized states. The predicted effective stress and effective strain fields in Fig. 10 reproduce the main regions of mechanical response observed in the ground truth.

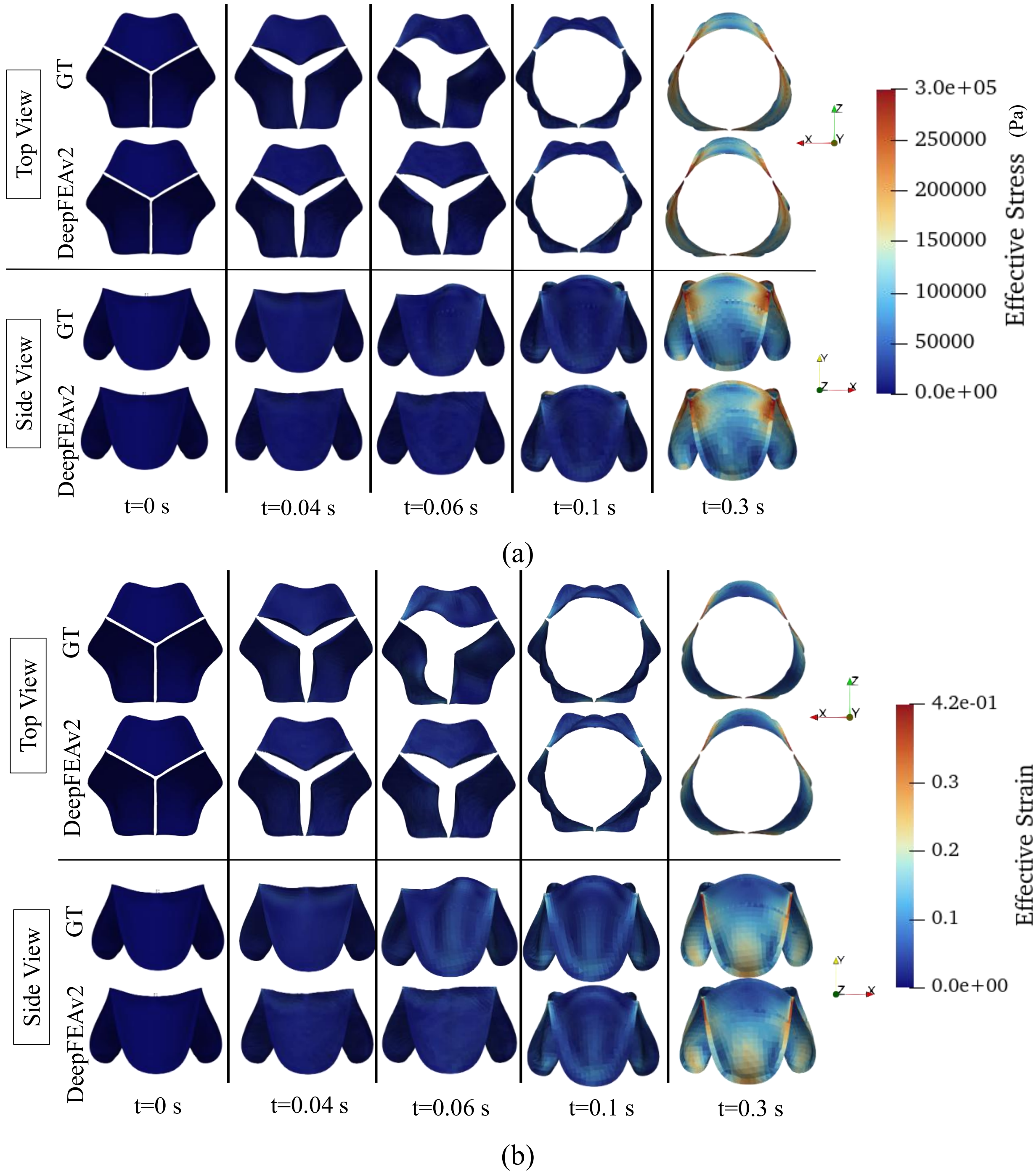

*Figure 10 Qualitative results for a representative simulation of the aortic valve dataset, depicting top view and side (a) effective stress in Pa and (b) effective strain view of the leaflet part, where GT denotes the ground truth.*

High-response areas appear in similar locations, particularly near the leaflet attachment and commissural regions, where adjacent leaflets meet and larger deformation gradients are expected during pressure loading. The agreement between the top and side views indicates that the model preserves the three-dimensional distribution of the mechanical fields, rather than matching the response only from a single projection.

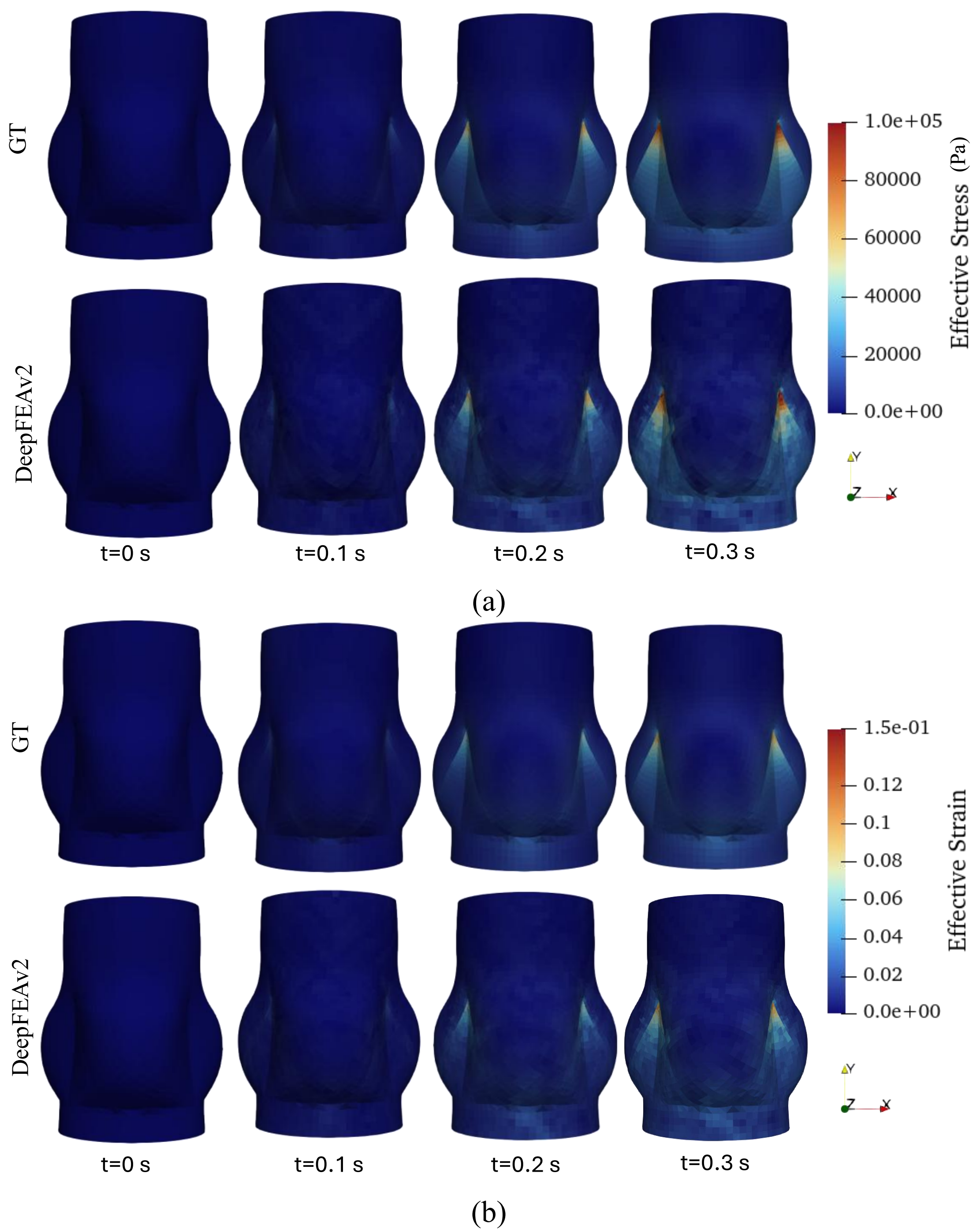

*Figure 11 Qualitative results depicting (a) effective stress in Pa and (b) effective strain for the root part of the simulation presented in Fig. 10, where GT denotes the ground truth.*

Figure 11 further evaluates the model on the aortic root using a side-view representation for the same simulation case as in Fig. 10. The predicted effective stress and strain fields follow the global spatial distribution of the ground truth and capture the progressive increase in mechanical response over time. The model reproduces the dominant response patterns of the surrounding root structure, including regions of increased stress and strain associated with larger local deformation. Although minor local discrepancies can be observed in some high-intensity regions, the overall temporal evolution and spatial organization of the mechanical fields are well preserved. Overall, the qualitative results in Figs. 10 and 11 support the quantitative findings and demonstrate the agreement of predictions across leaflet and root geometries, as well as across multiple viewing directions, suggesting that the proposed MCBG formulation can model transient, spatially heterogeneous mechanical behavior in anatomically complex FE structures.

The importance of these results lies in the biomechanical role of the aortic valve response. In this application, displacement prediction is directly related to the transient opening and deformation behavior of the valve leaflets under pressure loading, while effective stress and strain describe the local mechanical state of the valve tissue. Accurately predicting both quantities is therefore necessary to reproduce not only the global valve motion, but also the local tissue-level response of the leaflets and root. Moreover, preserving the spatial location of high-stress and high-strain regions is important because these regions indicate mechanically critical areas of the valve structure. The strong quantitative and qualitative agreement observed in this study therefore suggests that DeepFEAv2 can capture the key biomechanical features of a pressure-driven nonlinear shell simulation, supporting its potential use for more realistic valve modeling scenarios beyond the controlled solid linear elasticity datasets.

## 4.4 Inference Time Efficiency Assessment

The quantitative and qualitative results demonstrated the improved predictive accuracy of DeepFEAv2. However, the main motivation for using a deep learning-based surrogate model is to reduce the computational time required to obtain transient FEA responses. Therefore, a computational time analysis was performed by comparing the average time required to generate one complete transient simulation using conventional FEA with the average inference time required by DeepFEA and DeepFEAv2 to predict the same output sequence. The times were measured on a workstation equipped with an AMD Ryzen 5 7600X CPU and NVIDIA GeForce RTX 3090 GPU with 24 GB memory. The FEA time corresponds to the time required by LS-DYNA to complete one full transient simulation on the CPU. The surrogate models' average inference time corresponds to the time required to recursively predict the complete output sequence of one simulation.

*Table 4 Average CPU and GPU enabled inference time (in seconds) of DeepFEA and DeepFEAv2 compared to CPU enabled generation time of FEA for one simulation.*

| **Methods** | **Datasets** | | | | | |
|---|---|---|---|---|---|---|
| | Structured 3D LEM | | Unstructured 3D LEM | | Aortic Valve | |
| | *CPU (s)* | *GPU (s)* | *CPU (s)* | *GPU (s)* | *CPU (s)* | *GPU (s)* |
| *FEA* | 11.83 | N/A | 90.14 | N/A | 97.58 | N/A |
| *DeepFEA* | 0.84 | 0.07 | *N/A* | *N/A* | *N/A* | *N/A* |
| *DeepFEAv2* | 1.09 | 0.07 | 1.14 | 0.09 | 1.17 | 0.09 |

As shown in Table 4, DeepFEAv2 substantially reduced the computational time required to obtain complete transient responses compared with conventional FEA. On the Structured 3D LEM dataset, LS-DYNA required 11.83 s on average to generate one simulation, whereas DeepFEAv2 required 1.09 s on CPU and 0.07 s on GPU. This corresponds to speed-ups of 10.9× and 169.0×, respectively. Compared with the previous DeepFEA framework, DeepFEAv2 required a slightly higher CPU inference time on the structured dataset, increasing from 0.84 s to 1.09 s. This increase can be attributed to the increased complexity of the autoencoding module introduced in TopoNEP. However, both DeepFEA and DeepFEAv2 achieved the same GPU inference time of 0.07 s on this dataset. Therefore, DeepFEAv2 preserved the GPU-enabled inference efficiency of the previous framework while providing improved predictive accuracy and extending the model to mesh settings that DeepFEA cannot directly process.

For the Unstructured 3D LEM dataset, LS-DYNA required 90.14 s per simulation, whereas DeepFEAv2 required 1.14 s on CPU and 0.09 s on GPU. This corresponds to speed-ups of 79× and 1002×, respectively. Similarly, for the pressure-driven aortic valve dataset, LS-DYNA required 97.58 s per simulation, while DeepFEAv2 required 1.17 s on CPU and 0.09 s on GPU, corresponding to speed-ups of 83× and 1084×, respectively. These results demonstrate that DeepFEAv2 provides not only accurate prediction of transient FEA responses, but also practical computational acceleration. Therefore, the proposed framework can support rapid simulation-based assessment and repeated evaluation of different loading scenarios without requiring a full conventional FEA solver for each case.

# 5 Discussion

This study introduced DeepFEAv2, a deep-learning surrogate framework for transient FEA across different mesh topologies. The main objective was to overcome a key limitation of the previous DeepFEA framework, namely its dependence on regular grid-based mesh representations. This limitation is part of a broader challenge in deep learning-based FEA surrogate modeling: many existing architectures impose data structures that do not naturally match FE meshes. MLP-based models can approximate nonlinear relationships but generally ignore the spatial organization of the mesh, making them inefficient for large-scale NEO. CNN-based models improve spatial feature extraction but typically require regular grid-based input, which limits their applicability to complex unstructured meshes. Recurrent models such as LSTMs and GRUs can model temporal evolution, but they often operate on flattened inputs and therefore lose explicit spatial or topological information. GNNs provide a more natural representation of mesh connectivity, but repeated message-passing over large transient meshes can become computationally demanding, particularly during long recursive predictions. Physics-informed methods can improve physical consistency, but they are often tied to specific governing equations, material assumptions, and boundary conditions, reducing their flexibility across different simulation settings.

DeepFEAv2 was designed to address these limitations by using FE connectivity to organize nodal features into element-local groups before deep learning-based processing. Instead of converting the mesh into a regular grid-based representation or applying graph message passing at every timestep, the proposed framework uses the connectivity matrix to group nodal information into element-local groups. These groups are then sorted using adjacency-based ordering, compressed through an element-centric autoencoder, evolved in a compact latent space using 3D ConvLSTM layers, and decoded through dedicated node and element branches. This design allows DeepFEAv2 to preserve node-element coupling while remaining applicable to both structured and unstructured mesh topologies. A key strength of DeepFEAv2 is that it preserves the physical organization of FEA data while reducing the complexity of temporal modeling. The connectivity matrix provides a deterministic node-to-element mapping mechanism, while the element-centric autoencoder compresses the mesh representation before recurrent processing in the latent space. This allows the model to capture transient dynamics without directly processing the full FE mesh state at every timestep. The separate prediction branches also preserve the natural distinction between node-based and element-based output.

In addition, the experimental results demonstrate that DeepFEAv2 improves the predictive capability of the previous DeepFEA framework while substantially extending its range of applicability. On the Structured 3D LEM benchmark, where a direct comparison with DeepFEA is possible, DeepFEAv2 achieved substantially higher performance with up to a 38.0% relative increase in $R^2$ and up to an 87.1% reduction in normalized errors as reported in Table 1. This improvement is especially important for transient FEA because displacement is the output used to update the mesh configuration at subsequent timesteps. Therefore, the results on the structured dataset demonstrate that replacing the regular grid-based representation of DeepFEA with MCBG does not compromise but further improves predictive performance on regular meshes. Beyond the structured dataset results, the main advantage of DeepFEAv2 is its ability to operate on mesh topologies that cannot be directly processed by the previous DeepFEA framework. This capability was demonstrated on the Unstructured 3D LEM dataset, which contains a substantially larger and topologically irregular mesh compared with the structured dataset. As shown in Table 2, DeepFEAv2 achieved an $R^2$ of up to 0.97 and normalized errors as low as 1.42%. These results are important since they show that the proposed framework can process meshes of different sizes and organizations without imposing a regular Cartesian structure on the physical domain. Furthermore, the pressure-driven aortic valve application study further highlights the practical relevance of this this formulation for different FE mesh topologies. Unlike the LEM datasets, the aortic valve simulation involves a more complex shell-based geometry with hyperelastic material behavior and distributed ventricular pressure loading. These characteristics make the problem substantially different from force-driven solid-element benchmarks and its solution becomes unfeasible using the structured-grid approach utilized by DeepFEA. Despite this increased geometric and physical complexity, DeepFEAv2 achieved high predictive accuracy in this setting, with $R^2$ values over 0.98 and normalized errors lower than 1.51%, as reported in Table 3. Thus, the aortic valve study demonstrates that DeepFEAv2 can be applied to realistic pressure-driven shell-based biomechanical simulations, where accurate surrogate modeling requires resolving both global structural motion and localized mechanical response. In addition to improving predictive accuracy and mesh applicability, DeepFEAv2 substantially accelerates transient FEA prediction. As shown in Table 4, the proposed framework achieved up to a three-orders-of-magnitude reduction in computational time compared with conventional LS-DYNA simulations, while maintaining the same GPU inference time as the previous DeepFEA framework on the structured mesh dataset.

Overall, these results suggest that the principal contribution of DeepFEAv2 is threefold. First, it improves prediction performance over DeepFEA on the structured mesh simulations. Second, it removes the regular grid-based dependency of the previous framework, enabling application to unstructured meshes and complex mesh geometries. Third, it preserves the computational advantage expected from surrogate modeling, achieving up to three orders of magnitude inference acceleration compared with conventional FEA simulations. Compared with conventional CNN- or ConvLSTM-based surrogates, DeepFEAv2 avoids converting the physical mesh into a regular grid-based input representation, whereas when compared with purely graph-based methods, it uses deterministic FE connectivity and latent recurrent evolution rather than repeated full-graph message passing at each timestep. This positions DeepFEAv2 as a practical bridge between FE topology and spatiotemporal deep learning for transient mechanics prediction. Despite these merits, one limitation remains evident in the unstructured LEM results. The effective stress and strain predictions show lower accuracy than the displacement predictions. This suggests that element-based mechanical quantities are more difficult to recover on irregular meshes, particularly when localized stress or strain peaks occur over only a small number of elements. Therefore, future improvements should focus on strengthening the optimization strategy regarding element-based output in these types of simulations.

# 6 Conclusions

This study presented DeepFEAv2, a a deep learning surrogate model for transient FEA across different mesh topologies. The proposed framework extends the previous DeepFEA by replacing regular grid-based input representations with a MCBG–ABEO module and enabling FE topology to be preserved without imposing a structured grid on the physical mesh. The main conclusions of this study are summarized as follows:

- DeepFEAv2 provides a deep-learning surrogate framework that supports different FE mesh topologies without requiring regular grid-based representations. By using FE connectivity instead of regular grid-based representations, the framework can operate on both structured and unstructured FE meshes while preserving the physical node-element coupling defined by the mesh topology.
- The proposed TopoNEP network enables efficient recursive prediction over mesh-based simulation data of transient FEA. By combining a one-dimensional CNN autoencoder, 3D ConvLSTM layers, and dedicated node and element decoder branches, TopoNEP compresses the grouped element sequence produced by MCBG and ABEO into a compact latent space, models its transient evolution, and reconstructs NEO predictions for all nodes and elements of the original FE mesh without requiring the full high-dimensional representation to be temporally evolved directly at every timestep.

- DeepFEAv2 supports simultaneous prediction of NEO through a FEA-informed node-element optimization strategy. The framework jointly predicts nodal displacement, effective stress, and effective strain, while the extended loss formulation incorporates geometry- and time-consistency terms to improve the accuracy of the predicted output.
- The experimental evaluation demonstrates that DeepFEAv2 improves upon the previous DeepFEA framework on structured mesh simulations and extends transient prediction to unstructured meshes, including pressure-driven shell-based biomechanical simulations, achieving $R^2$ values up to 0.99, normalized errors as low as 0.38%, and substantial computational acceleration of up to three orders of magnitude compared with conventional FEA simulations.

These findings indicate that DeepFEAv2 provides a flexible and scalable surrogate modeling approach for transient FEA simulations, capable of supporting increasingly complex FE settings across different mesh types, element formulations, and loading mechanisms. As a result, the proposed framework can support faster simulation-based assessment, enable rapid exploration of loading or design scenarios, and provide a foundation for future analysis workflows without requiring repeated FEA simulation runs. Despite these advantages, element-based effective stress and strain prediction remains a key direction for further improvement, particularly for irregular meshes where localized mechanical peaks may occur over a small number of elements. Thus, future work will focus on improving local stress and strain accuracy and evaluating DeepFEAv2's performance on more diverse real-world engineering and biomechanical datasets.